\documentclass[journal]{IEEEtran}

\usepackage[pdftex]{graphicx}
\usepackage{amsopn}
\usepackage{booktabs} % For formal tables
\usepackage{url}
\usepackage{color}
\usepackage{multirow}
\usepackage{color}
\usepackage[switch]{lineno}
\usepackage{graphicx}
\usepackage{subfigure}
\usepackage{url}
\usepackage{epstopdf}
\usepackage{hyperref}
\usepackage[table,xcdraw]{xcolor}
\usepackage{amsmath}
\usepackage{textcomp}
\usepackage{verbatim}
\usepackage{bm}

\usepackage{amssymb}
\usepackage{amsthm}
\usepackage{amsfonts}
\usepackage{enumitem}
\usepackage{comment}
\usepackage{amssymb}
\usepackage{multirow}

\theoremstyle{definition}

\ifCLASSOPTIONcompsoc
  \usepackage[nocompress]{cite}
\else
  \usepackage{cite}
\fi

\ifCLASSINFOpdf
\else
\fi

\begin{document}

\title{Road Network Guided Fine-Grained Urban Traffic Flow Inference}

\author{Lingbo Liu,
        Mengmeng Liu,
        Guanbin Li,
        Ziyi Wu,
        Junfan Lin,
        and Liang Lin, {\textit{Senior Member, IEEE}}

\IEEEcompsocitemizethanks{
\IEEEcompsocthanksitem This work was supported in part by the National Natural Science Foundation of China (No.62306258, No.62322608, No.61976250, No.62272494, No.61836012 and U1811463). Lingbo Liu and Mengmeng Liu are co-first authors. (Corresponding author: Guanbin Li.)
\IEEEcompsocthanksitem All authors are with the School of Computer Science and Engineering, Sun Yat-Sen University, China, 510000.
\IEEEcompsocthanksitem L. Liu is also with Department of Land Surveying and Geo-Informatics, The Hong Kong Polytechnic University, Hong Kong, China.
}}

%\markboth{IEEE Transactions on Neural Networks and Learning Systems}%
%{Liu \MakeLowercase{\textit{et al.}}: Road Network Guided Fine-Grained Urban Traffic Flow Inference}

\IEEEtitleabstractindextext{%
\begin{abstract}
  Accurate inference of fine-grained traffic flow from coarse-grained one is an emerging yet crucial problem, which can help greatly reduce the number of the required traffic monitoring sensors for cost savings. In this work, we notice that traffic flow has a high correlation with road network, which was either completely ignored or simply treated as an external factor in previous works.
  To facilitate this problem, we propose a novel Road-Aware Traffic Flow Magnifier (RATFM) that explicitly exploits the prior knowledge of road networks to fully learn the road-aware spatial distribution of fine-grained traffic flow. Specifically, a multi-directional 1D convolutional layer is first introduced to extract the semantic feature of the road network. Subsequently, we incorporate the road network feature and coarse-grained flow feature to regularize the short-range spatial distribution modeling of road-relative traffic flow. Furthermore, we take the road network feature as a query to capture the long-range spatial distribution of traffic flow with a transformer architecture. Benefiting from the road-aware inference mechanism, our method can generate high-quality fine-grained traffic flow maps. Extensive experiments on three real-world datasets show that the proposed RATFM outperforms state-of-the-art models under various scenarios. Our code and datasets are released at {\color{blue}{\url{https://github.com/luimoli/RATFM}}}.
\end{abstract}

}

\maketitle

\IEEEdisplaynontitleabstractindextext
\IEEEpeerreviewmaketitle

%\IEEEraisesectionheading{\section{Introduction}\label{sec:introduction}}
\IEEEpeerreviewmaketitle
\section{Introduction}\label{sec:introduction}

\IEEEPARstart{I}{n} the past decades, a large number of people have poured into the cities, and the number of vehicles has increased sharply, which brings great challenges to urban transportation management \cite{zheng2014urban,lou2020probabilistic}. In the construction of intelligent transportation systems (ITS), traffic flow monitoring is a crucial component that provides significant information for traffic congestion warnings and traffic state prediction. To obtain citywide fine-grained traffic flow data (e.g., inflow, outflow, volume), transportation departments usually deploy large amounts of sensing devices (e.g., surveillance cameras and induction loops) to cover each small geographical region. However, it is a heavy economic burden to purchase these devices. For instance, it was reported that the ITS project of Hong Kong spent around 78 million USD on monitoring equipment expenditure \cite{HKITS}. Moreover, as time goes by, a lot of equipment will be depreciated and updated, which also requires huge expenses. Under these circumstances, we desire a novel technology that can greatly reduce the number of deployed sensors while effectively maintaining the fine granularity of traffic flow information.

\begin{figure*}[t]
  \centering
  \includegraphics[width=0.85\linewidth]{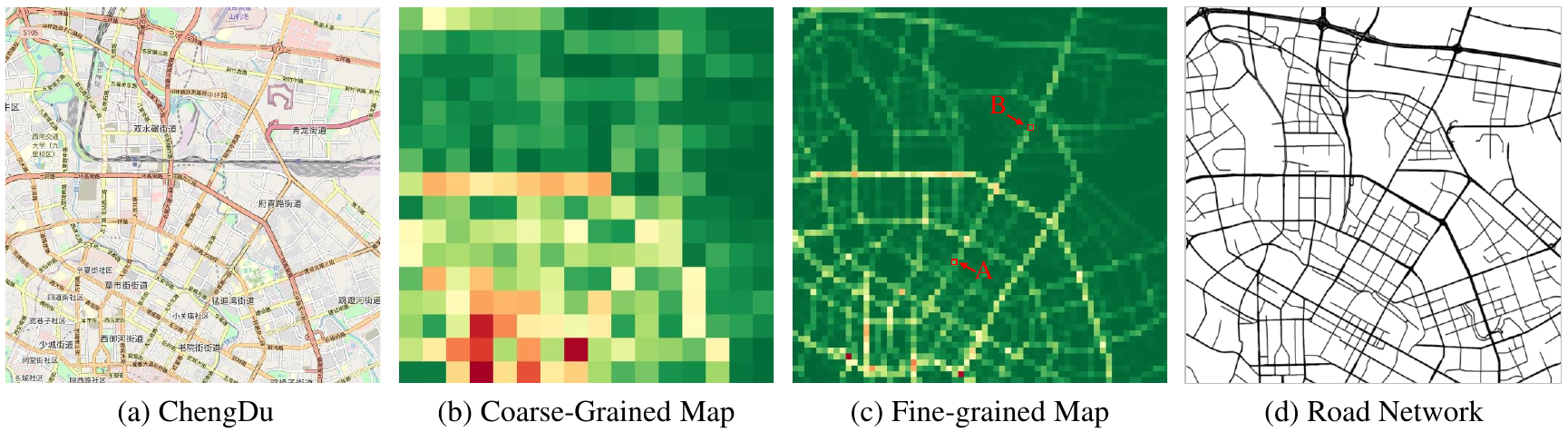}
  \vspace{-3mm}
  \caption{Illustration of (b) coarse-grained traffic flow map, (c) fine-grained traffic flow map, and (d) road network map of the studied city (e.g., ChengDu, China). We notice that traffic flow has a high correlation with traffic road network, i,e, road regions usually have high traffic flow. Therefore, the road network can be taken as prior knowledge to facilitate the fine-grained traffic flow inference.}
  \label{fig:intro}
\end{figure*}

In this work, we focus on how to accurately generate fine-grained data from coarse-grained data collected with a small number of traffic sensors, which is termed fine-grained urban traffic flow inference. Due to its great social benefits, this problem has recently received increasing attention in both industry and academic communities. In most preliminary works~\cite{zong2019deepdpm,liang2019urbanfm,ouyang2020fine}, the city being studied is first divided into a coarse grid map and a fine grid map on the basis of latitude and longitude coordinates, as shown in Fig. \ref{fig:intro}-(b,c). Note that the traffic flow observed on the coarse grid map is termed a coarse-grained traffic flow map. Likewise, the fine-grained traffic flow map is observed on the fine grid map. Inspired by the success of deep learning \cite{goodfellow2016deep,bengio2017deep}, conventional methods usually employ convolutional neural networks (CNN) to learn mapping functions between coarse-grained traffic flow maps and fine-grained ones. Following \cite{sun2020predicting,guo2021learning}, some external factors (e.g., weather and timestamp) are incorporated to learn the fine-grained traffic flow distribution.

Despite recent progress, fine-grained traffic flow inference remains a challenging task. {\bf{First}}, different from image super-resolution where low-resolution images still have obvious structures \cite{tang2019deep,xin2020wavelet}, our coarse-grained traffic flow maps are relatively rough, without any structural information, as shown in Fig. \ref{fig:intro}-(b). It is very difficult to directly transform coarse-grained maps to fine-grained maps. Fortunately, we observe that the urban road network has an obvious structure that is perfectly aligned with the fine-grained traffic flow distribution, as shown in Fig. \ref{fig:intro}-(d), since most vehicles drive on these traffic roads. In this case, the road network can be regarded as an instructive prior knowledge for traffic flow inference. Nevertheless, most previous methods \cite{zong2019deepdpm,liang2019urbanfm,ouyang2020fine} were not aware of this knowledge.
{\bf{Second}}, how to model the road network is still an open problem. We can see that blank background dominates the road network map, while those traffic roads are thin and long. We argue that it is suboptimal to model the road network with 2D convolutional layers commonly used in image analysis since the square filters would indulge in the dominant background and belittle those traffic roads. Thus, a more effective method is desired to capture the road network distribution.
{\bf{Third}}, despite short-range spatial dependencies were widely explored in \cite{zong2019deepdpm,liang2019urbanfm,ouyang2020fine}, we argue that long-range spatial dependencies are also crucial for traffic flow inference. For instance, two distant regions may have similar or correlative traffic flow, if they belong to the same traffic road, such as the regions A and B in Fig. \ref{fig:intro}-(c). Therefore, it is meaningful to capture the traffic flow long-range dependencies, especially for traffic road regions.

Taking into consideration the above challenges, we propose a novel Road-Aware Traffic Flow Magnifier (RATFM), which fully exploits the prior knowledge of road network to learn the distribution patterns of fine-grained traffic flow. Specifically, our RATFM is composed of a road network modeling module, a short-range road-aware inference module, a long-range road-aware inference module, and an external factor modeling module.
First, we crawl the traffic network map of the studied city from Internet, and introduce a unified multi-directional 1D convolutional layer to effectively model traffic roads with four groups of 1D filters in various directions.  Then, the road network feature and the coarse-grained traffic flow feature are incorporated and fed into stacked residual blocks \cite{he2016deep} to better model the short-range spatial distribution of road-relative traffic flow. Moreover, we take the road network feature as query priori and capture the long-range spatial distribution of traffic flow using a global attention transformer \cite{carion2020end}. Thanks to the tailor-designed road-aware mechanism, our method is capable to generate high-quality traffic flow maps with a spatial fine-granularity. We conduct extensive experiments on three real-world benchmarks, and the evaluation results show that the proposed RATFM outperforms existing state-of-the-art methods under various scenarios.

In summary, our major contributions of this paper are summarized in the following aspects:
\begin{itemize}
  \item We re-examined the problem of fine-grained traffic flow inference from the perspective of urban road network relevance, and extend the existing pure data-driven representation learning to knowledge-guided representation learning. Specifically, we propose a novel Road-Aware Traffic Flow Magnifier, which takes the road network as prior guidance to effectively learn the fine-grained distribution of urban traffic flow.

  \item For road network modeling, a multi-directional 1D convolutional layer is specially introduced to better capture the distribution patterns of those spindly traffic roads and alleviate the distraction of background.
  \item We introduce a transformer-based long-range inference module, which takes the road network knowledge as a query to jointly reason the traffic flow distribution of all regions with fine granularity.
    \item Extensive experiments on three real-world benchmarks show the effectiveness of the proposed RATFM for fine-grained urban traffic flow inference.
\end{itemize}

The rest of this paper is organized as follows. First, we review some related works of traffic flow analysis in Section \ref{sec:review}. We then introduce the proposed Road-Aware Traffic Flow Magnifier in Section \ref{sec:method}. To verify the effectiveness of RATFM, we perform extensive experiments conducted in Section~\ref{sec:experiment}, and conclude this paper in Section~\ref{sec:conclusion}.

\section{Related Work}\label{sec:review}
\subsection{Traffic Flow Analysis}
Traffic flow analysis is crucial for intelligent transportation systems, since the analysis results can provide important information for various downstream applications. Over the past decades, massive efforts \cite{liu2018attentive,liang2018geoman,guo2019deep,song2020spatial,bai2020adaptive,chen2021bayesian} have been made to address the traffic analysis problem. For instance, Han {\it{et al.}} \cite{han2020traffic} developed a Filter-Discovery-Match framework to learn incident patterns from vehicle trajectories for traffic incident detection, while Sun {\it{et al.}} \cite{sun2020predicting} employed spatial graph convolution to build a multi-view graph convolutional network for crowd flow prediction.

Recently, fine-grained traffic flow inference has become an emerging problem, due to its huge potential for device cost savings. In the literature, some deep learning-based methods have been proposed for this task. For instance, Zong {\it{et al.}} \cite{zong2019deepdpm} utilized a super-resolution convolutional neural network to learn the spatial mapping between coarse-grained and fine-grained traffic flow maps. Liang {\it{et al.}} \cite{liang2019urbanfm} incorporated a deep residual network and a distributional upsampling module to generate fine-grained flow distributions. Ouyang {\it{et al.}} \cite{ouyang2020fine} employed a cascade pyramid network called UrbanPy to upsample the coarse-grained inputs progressively. Zhou {\it{et al.}} \cite{zhou2021inferring} proposed a deep neural network called UrbanODE, which incorporated Neural Ordinary Differential Equations \cite{chen2018neural} and pyramid attention to learn spatial correlations of urban traffic flow in surrounding regions. Yu {\it{et al.}} \cite{yu2023Overcoming} extracted spatial features from both local and global levels, and combined temporal features to generate the fine-gained maps using time-specific convolutional layers.

The most relevant work to ours is UrbanPy \cite{ouyang2020fine}, which also uses road network structures. However, there are two key differences between our RATFM and UrbanPy in terms of road network usage. {\bf{Firstly}}, UrbanPy only uses the statistic road density/number for each region, missing a lot of crucial information of road network structures, e.g., road direction and connectivity. By contrast, our RATFM learns the road distribution patterns effectively from high-resolution road network maps using a tailor-designed multi-directional 1D convolutional layer. {\bf{Secondly}}, UrbanPy directly takes the concatenation of road density features and traffic flow features for representation learning, which can't well model the correlation between road network and traffic flow. By contrast, our RATFM introduces two short and long-range road-aware inference modules to effectively generate fine-grained traffic flow maps. Thanks to these tailor-designed modules, our method can fully exploit the prior knowledge of urban road networks to facilitate the problem of fine-grained traffic flow inference.

\subsection{Image Super-Resolution Reconstruction}
Image super-resolution reconstruction \cite{cao2016image} is a classical problem that aims at generating high-resolution images from low-resolution ones. Numerous deep learning-based models have been proposed for this task. As a pioneering work, Dong {\it{et al.}} \cite{dong2014learning} developed the first Convolutional Neural Network based method for super-resolution, which is composed of three convolutional layers to reconstruct high-resolution images. Tai {\it{et al.}} \cite{Tai_2017_CVPR} proposed a Deep Recursive Residual Network with 52 convolutional layers to learn the residual information between low/high-resolution images. Zhang {\it{et al.}} \cite{zhang2016coarse} developed a coarse-to-fine super-resolution recovery framework, which is accomplished by the use of a CNR-based algorithm for detail synthesis and followed by a nonlocal structural regression regularization algorithm for quality enhancement. Recently, temporal information has also been explored in this field. For instance, Kim {\it{et al.}} \cite{liang2020video} incorporated a deep convolutional neural network and a spatial-temporal feature similarity calculation method to learn the nonlinear correlation mapping between low-resolution and high-resolution video frame patches. Despite progress, these methods of image super-resolution reconstruction can not be directly applied to the more challenging fine-grained traffic flow inference, where coarse-grained traffic flow maps are too rough to maintain the structural information. Base on this concern, we incorporate the structural information of road networks to guide the fine-grained traffic flow inference.

\begin{figure*}[ht]
  \centering
  \includegraphics[width=\linewidth]{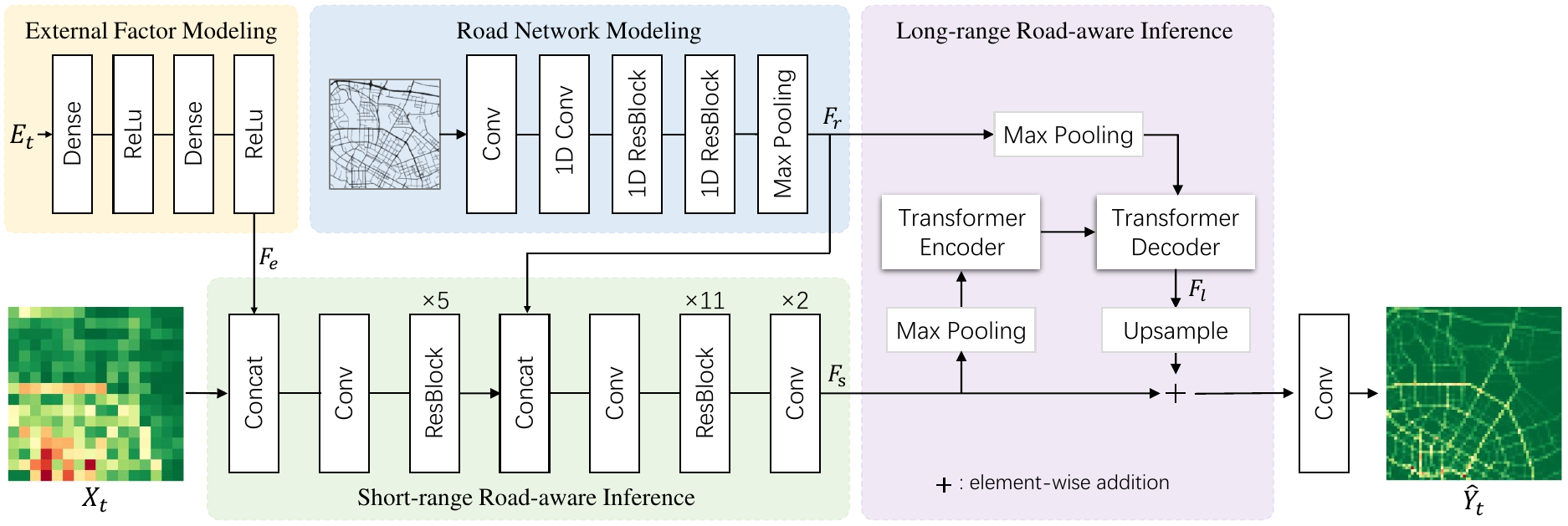}
  \caption{The architecture of the proposed Road-aware Traffic Flow Magnifier for fine-grained urban traffic flow inference. Specifically, our framework is composed of a road network modeling module, a short-range road-aware inference module, a long-range road-aware inference module, and an external factor modeling module. The coarse-grained map $X_{t} \in \mathbb{R}^{I_c{\times}J_c{\times}S}$ is first resized to the resolution $I_f{\times}J_f$ and then fed into our network. $F_e \in \mathbb{R}^{I_f{\times}J_f{\times}1}$ is the extracted feature of the external factor $E_t$. The learnable representation of the input road network is denoted as $F_r \in \mathbb{R}^{I_f{\times}J_f{\times}C}$, where $C$ is the number of channels. $F_s \in \mathbb{R}^{I_f{\times}J_f{\times}C}$ is the short-range road-aware feature, while $\frac{I_f}{2}{\times}\frac{J_f}{2}{\times}C$ is the short-range road-aware feature and it would be resized to the resolution $I_f{\times}J_f$.}
  \label{frame}
\end{figure*}

\subsection{Transformer Architecture}

Transformer \cite{vaswani2017attention} is an advanced neural network block that aggregates information from the entire input sequence with an attention mechanism \cite{bahdanau2014neural}. Specifically, the transformer is composed of self-attention layers, a point-wise feed-forward layer, and layer normalization. The main advantage of
the transformer is its global computation and perfect memory mechanism, which makes it more suitable than Recurrent Neural Networks on long sequences. Nowadays, transformer is widely studied in various tasks, including natural language processing \cite{devlin2018bert,radford2019language}, computer vision \cite{xu2021multigraph,li2021groupformer} and data mining \cite{liu2021online}. Inspired by the success of these works, we apply a transformer to learn the long-range spatial distribution of fine-grained traffic flow. However, unlike previous works whose decoders take position information as query, our method adopts the road network prior as query, making our network better learn the fine-grained flow in road regions. To the best of our knowledge, our work is the first attempt to employ attention-based transformer to address fine-grained traffic flow inference.

%----------------------------------------------------------------------------------------------------------------------------------
\section{Methodology} \label{sec:method}

In this section, we propose a novel Road-aware Traffic Flow Magnifier (RATFM), which can generate fine-grained traffic flow maps accurately under the guidance of road network information. As shown in Figure \ref{frame}, our RATFM consists of four components, including {\bf\color{red}{i)}} a road network modeling module, {\bf\color{red}{ii)}} a short-range road-aware inference module, {\bf\color{red}{ii)}} a long-range road-aware inference module and {\bf\color{red}{iv)}} an external factor modeling module.

Before introducing the details of RATFM, we define some notations of fine-grained traffic flow inference. In this work, we do not adopt the graph-structure strategy \cite{sun2020predicting} for urban region partition, because graph-based traffic flow maps are not aligned with the Euclidean-based road network map. Following previous works \cite{liu2019contextualized,ouyang2020fine}, we partition the studied city into a regular grid map based on the coordinates of latitude and longitude, as shown in Fig \ref{fig:intro}. It is worth noting that the granularity of flow maps is related to the partition setting. More specifically, coarse-grained maps are observed upon a small number of grids, each of which denotes a large geographic area in the real world. Therefore, the resolution of coarse-grained maps is set to a tuple $I_c{\times}J_c$ with small values.
For convenience, the citywide coarse-grained traffic flow map at time interval $t$ is represented as ${X_{t}} = \left[ x_{t}^{\left(1,1\right)},\dots,x_{t}^{\left(i,j\right)},\dots,x_{t}^{\left(I_c,J_c\right)} \right] \in \mathbb{R}^{I_c \times J_c \times S}$, where $x_{t}^{\left(i,j\right)}\in \mathbb{R}^S$ records the $S$ traffic flow states (e.g., vehicle volume, inflow and outflow) of the grid $\left(i,j\right)$.

Intuitively, the granularity of fine-grained traffic flow maps should be more refined than that of coarse-grained traffic flow maps. To this end, each grid of coarse-grained maps is further divided into $N{\times}N$ grids with a small geographic acreage. In this case, the resolution of fine-grained maps is $I_f{\times}J_f = NI_c{\times}NJ_c$.
Given a coarse-grained traffic flow map ${X_t}$, our goal is to generate a fine-grained traffic flow map
$Y_{t} = \left[ y_{t}^{\left(1,1\right)},\dots,y_{t}^{\left(i,j\right)},\dots,y_{t}^{\left(I_f, J_f\right)} \right] \in \mathbb{R}^{I_f \times J_f \times S}$.

\begin{table}[t]
 \caption{Some notations for fine-grained traffic flow inference.}
  \vspace{-2mm}
  \newcommand{\tabincell}[2]{\begin{tabular}{@{}#1@{}}#2\end{tabular}}
  \centering
  \resizebox{9cm}{!} {
    \begin{tabular}{l|l}
    \hline
    {\textbf{Notations}} & {\textbf{Description}} \\
    \hline
    \hline
    $I_c{\times}J_c$ & the resolution of coarse-grained traffic flow maps\\
    \hline
    $I_f{\times}J_f$ & \tabincell{l}{the resolution of fine-grained traffic flow maps\\($I_f{\times}J_f$ is much larger than $I_c{\times}J_c$)}\\
    \hline
    $S$ & the number of the studied traffic state variables\\
    \hline
    $X_{t} \in \mathbb{R}^{I_c{\times}J_c{\times}S}$ & the coarse-grained flow map at time interval $t$\\
    \hline
    $Y_{t}{~}\in \mathbb{R}^{I_f{\times}J_f{\times}S}$ & the fine-grained flow map at time interval $t$\\
    \hline
    $M_r \in \mathbb{R}^{2I_f{\times}2J_f}$ & the resized road network map\\
    \hline
    $E_t$ & the external factor vector at time interval $t$\\
    \hline
    \end{tabular}
  }
  \label{tab:Notations}
\end{table}

\subsection{Road Network Modeling Module} \label{sec:road_modeling}
As mentioned above, the distribution of fine-grained traffic flow usually has remarkable relevance to urban road networks. In this subsection, we describe how to construct the road network map of the studied city and extract the deep semantic feature of the road network, which is used as prior knowledge to guide the fine-grained traffic flow inference in the following sections.

\subsubsection{Road Network Generation} \label{sec:road_network_generation}
In this work, we crawl the traffic road information of the studied city from OpenStreetMap (OSM \cite{osm}). On this website, 27 types of roads are recorded and we can obtain the detailed category of each road. To better fit the traffic flow distribution, we retain those primary traffic roads on which vehicles tend to drive, such as highways, secondary roads, trunks, etc, and discard some infrequently or impossibly traveled roads like paths, railways, cycleways, etc.
With $Arcmap$ as an auxiliary tool\footnote{\url{https://desktop.arcgis.com/zh-cn/arcmap/}}, these retained traffic roads are then rendered to construct a road network map. Notice that the width and shape of roads may vary in this generated map. For example, those roads with higher traffic levels usually appear as wider lines, while those large-scale roundabouts may appear as circles, as shown in Fig. \ref{fig:intro}-(d).

For further processing, the rendered road network map is organized into a grey-scale image, where the value at each pixel indicates the existence and intensity of traffic roads at the corresponding geographic area. In general, road-related pixels usually have high values, while the other pixels always have a value of zero. Notice that the original resolution of the road network map is much higher than that of the fine-grained traffic flow maps. To facilitate the representation extraction, the road network map is downsampled to $2I_f{\times}2J_f$ with bilinear interpolation. Such a resized road map is denoted as $\overline{M}_{r} \in \mathbb{R}^{2I_f{\times}2J_f}$.

Nevertheless, we find that $\overline{M}_{r}$ still suffers from some flaws. For instance, compared with the downtown roads, some suburban highways and trunks have less traffic flow but higher intensity values on the road network map, as shown in Fig. \ref{fig:intro}. To eliminate this issue, we assign adaptive weight to each road on the basis of historical flow. Specifically, we first generate a weight map $M_w \in \mathbb{R}^{2I_f{\times}2J_f}$ by computing the average of historical coarse-grained flow maps and resizing the average map to $2I_f{\times}2J_f$ with nearest-neighbor interpolation. The final road network map $M_r \in \mathbb{R}^{2I_f{\times}2J_f}$ is calculated as:
\begin{equation}
  M_r = \overline{M}_r \otimes  M_w,
\end{equation}
where $\otimes$ denotes an element-wise multiplication.

\subsubsection{Road Network Representation Learning}
We then transfer the generated road network map into a high-dimensional representation with a convolutional neural network.
In image processing field \cite{liu2019facial,liu2019contextualized,liu2020efficient,liu2020dynamic}, convolutional operators usually use square kernels (e.g.,$3\times 3$) as main components to model natural objects with clear boundaries and shapes. However, traffic roads are thin and long on the map, significantly different from natural objects. It is suboptimal to model traffic roads with square kernels, since the road-related pixels are sparse and many background pixels are involved. Fortunately, we observe that 1D filters are more aligned with road shapes. Inspired by this finding, we develop a multi-directional 1D convolutional neural network with 1D filters to effectively learn the feature representations for the urban road network.

\begin{figure}[t]
  \centering
  \includegraphics[width=0.9\linewidth]{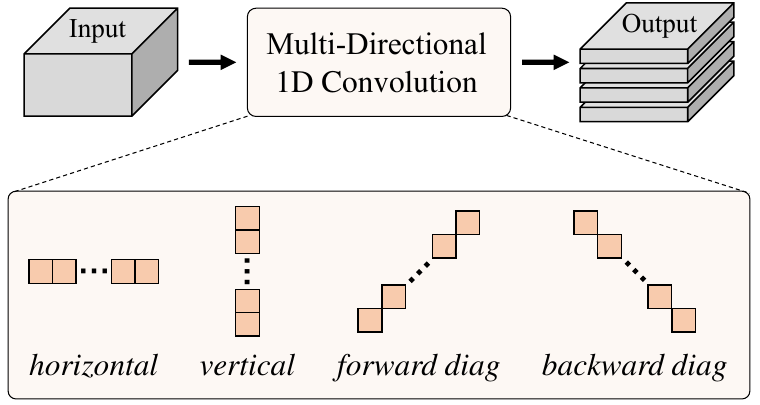}
  \vspace{-2mm}
  \caption{Illustration of the proposed multi-directional 1D convolutional layer. In this layer, four groups of 1D convolutional filters are utilized to model the traffic roads of different directions, including horizontal, vertical, forward diagonal, and backward diagonal ones.}
  \label{1D_Conv}
\end{figure}

As shown in Fig. \ref{1D_Conv}, the proposed multi-directional 1D convolutional layer contains four groups of 1D filters, which are utilized to model the traffic roads of different directions, including horizontal, vertical, forward diagonal, and backward diagonal ones. For convenience, the input of 1D convolution is denoted as $F \in \mathbb{R}^{H{\times}W{\times}C}$, where $H$ and $W$ are height and width, $C$ is the number of channels. Notice that each direction is composed of $\frac{C}{4}$ filters. More specifically, the 1D filters of the direction $d$ is denoted as $K_d \in \mathbb{R}^{C{\times}(2R+1){\times}\frac{C}{4}}$, where $R$ is the radius of kernels. By definition, the output feature $\hat{F_d} \in \mathbb{R}^{H{\times}W{\times}\frac{C}{4}}$ of the direction $d$ is computed by:

\begin{small}
\begin{equation}
  \hat{F_d}\left [h,w,c \right ] =\sum_{i=-R}^{R}\sum_{j=0}^{C-1} {F_d}\left [ h+i\cdot{I_d^h},w+i\cdot{I_d^w},j \right ]\cdot K_d\left [j,i+R,c \right ] \notag
\end{equation}
\end{small}%
where $\hat{F_d}\left [h,w,c \right ]$ is the responded value at location $(h, w)$ for the $c$-th filter. $I_d = (I_d^h,I_d^w)$ is the direction indicator vector and set to $(0,1), (1,0),  (1,1), (1,-1)$ for horizontal, vertical, forward diagonal, and backward diagonal convolution, respectively. Finally, the outputs of all directions are concatenated to generate a comprehensive feature $\hat{F} \in \mathbb{R}^{H{\times}W{\times}C}$, which can better capture the information of linelike roads of various directions.

\begin{figure}[t]
  \centering
  \includegraphics[width=0.975\linewidth]{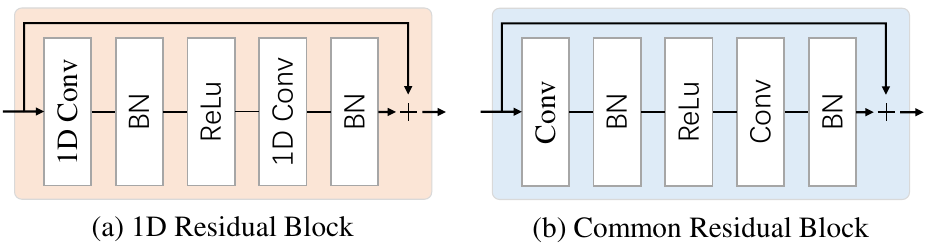}
  \vspace{-2mm}
  \caption{Illustration of our 1D residual block and common residual block. ``+'' denotes an element-wise addition operator.}
  \label{fig:ResBlock}
\end{figure}

Based on the multi-directional 1D convolutional layers, we develop a unified road network representation learning branch, whose architecture is shown in the blue block of Fig. \ref{frame}. First, the weighted road map $M_r$ is fed into a common convolutional layer for initial feature extraction. We then employ a 1D convolutional layer and two 1D residual blocks \cite{he2016deep} to generate a road-aware feature. As shown in Fig. \ref{fig:ResBlock}-(a), our 1D residual block contains two 1D convolutional layers followed by two Batch Normalization and a ReLU function. To enlarge the receptive field, the kernel size of the proposed 1D convolutional layers is set to a big value (e.g., 9) in our work. Finally, we adjust the resolution of the output feature using a $2\times2$ max-pooling layer. For convenience, the final road network feature is denoted as $F_r \in \mathbb{R}^{I_f{\times}J_f{\times}C}$. It is worth noting that each neuron on $F_r$ primarily captures the distribution patterns of those traffic roads with predefined directions in its receptive field, while also capturing the distribution patterns of those irregularly-shaped traffic roads to a certain extent. This is consistent with the actual distribution of urban traffic road networks, where most roads are horizontal/vertical or backward, and few roads are irregularly shaped.

\subsection{External Factor Modeling Module}
As mentioned in previous works \cite{liu2018attentive,liu2019contextualized,liu2020dynamic,ouyang2020fine}, traffic flow is usually affected by various external factors, such as meteorological factors and time factors. Hence we develop an external factor modeling module to model the influence of those factors, whose semantic feature would be incorporated to estimate the fine-grained traffic flow.

In this work, we collect various meteorological factors (i.e., weather condition, temperature, and wind speed) from the public website Wunderground\footnote{\url{https://www.wunderground.com/}}. Specifically, the weather condition is categorized into 10+ categories (e.g., sunny and rainy) and each category is digitized as an ordinal value, while temperature and wind speed are scaled into the range [0, 1] with a min-max linear normalization. Meanwhile, some time factors, i.e., day of the week and time of the day, are also collected and transformed into ordinal values. Finally, all digitized factors are concatenated to generate an external factor vector $E_t$.
As shown in Fig. \ref{frame}, the external factor vector $E_t$ is then fed into a Multi-layer Perceptron, which is implemented with two dense layers followed by ReLU functions. Specifically, these dense layers have 128 and one output neurons, respectively. The output of the second dense layer is then copied $I_f{\cdot}J_f$ times because the external factors are shared for all regions. Thus, the final external factor feature can be represented as $F_{e} \in \mathbb{R}^{I_f{\times}J_f{\times}1}$.

\subsection{Short-range Road-aware Inference Module}
Generally, the traffic flow at a fine-grained region is usually relevant to that of the corresponding coarse-grained region, thus capturing the short-range spatial dependencies is crucial \cite{zong2019deepdpm,ouyang2020fine,liang2021fine}. In this work, we find that the fine-grained traffic flow is also affected by the road network structure. Intuitively, road-related regions usually have high flow volume, while background regions have low flow volume. Based on these observations, we develop a Short-range Road-aware Inference Module, which employs local convolutional operations to model the short-range spatial dependencies of fine-grained traffic flow under the guidance of the traffic road network.

Specifically, the coarse-grained traffic flow map $X_t$ is first resized to $I_f{\times}J_f$ with bilinear interpolation. The upsampled coarse map and the external factor feature $F_{e}$ are concatenated and fed into a $9{\times}9$ convolutional layer for heterogeneous information fusion. The fused feature is then fed into five common residual blocks to learn the local spatial dependencies of traffic flow. As shown in Fig. \ref{fig:ResBlock}-(b), each residual block is composed of two $3{\times}3$ convolutional layers, two Batch Normalization, and a ReLU function. The output of the fifth residual block is denoted as $F_i \in \mathbb{R}^{I_f{\times}J_f{\times}C}$. Then, we introduce the traffic road network as prior knowledge to guide the inference network to
focus on road-related regions. Here the intermediate feature $F_i$ and the road network feature $F_r$ are concatenated and fused with a convolutional layer. The output of this convolutional layer contains the information of the road network and is fed into 11 residual blocks to effectively learn the fine-grained traffic flow of road-related regions. The output of the last residual block is further embedded to generate the final short-range traffic flow feature $F_s \in \mathbb{R}^{I_f{\times}J_f{\times}C}$ with two convolutional layers. By introducing the road network information, our feature $F_s$ is capable to better learn the local spatial distribution of fine-grained traffic flow.

\subsection{Long-range Road-aware Inference Module} \label{sec:long_range}
In addition to short-range dependencies, we observe that long-range spatial dependencies are also helpful for fine-grained traffic flow inference. For instance, two regions of the same road usually have a high correlation in terms of traffic flow, even if they are far apart in distance. However, previous CNN-based methods fail to effectively capture the long-range dependencies, since the effective receptive field of CNN occupies only a fraction of its full theoretical receptive field \cite{luo2016understanding}. Therefore, in this subsection, we develop a long-range road-aware inference module, which employs an advanced transformer \cite{vaswani2017attention,carion2020end} equipped with road prior to fully model the flow relations between all spatial regions.

By definition, a transformer is composed of an encoder and a decoder. As shown in Fig. \ref{frame}, we first adopt the standard encoder to generate an initial global feature. Specifically, to reduce the computational cost, the short-range feature $F_s$ is first downsampled to $\frac{I_f}{2}{\times}\frac{J_f}{2}$ with a $2{\times}2$ max-pooling layer.
The downsampled feature, denoted as $F_s^{2\downarrow}$, is then reshaped to $\frac{{I_f}{J_f}}{4}{\times}C$ and fed into the transformer encoder that mainly contains a self-attention unit and a feed-forward network. The formulation of our encoder can be simply expressed as:
\begin{equation}
\begin{split}
Q_e &= F_s^{2\downarrow}*\mathcal{W}_e^q,{~~}K_e = F_s^{2\downarrow}*\mathcal{W}_e^k,{~~}V_e = F_s^{2\downarrow}*\mathcal{W}_e^v, \\
A_e &= F_s^{2\downarrow}+softmax(\frac{Q_eK_e}{\sqrt{C}})V_e, \\
F_l^e &= A_e + F_e(A_e), \\
\end{split}
\end{equation}
where the parameter $Q_e, K_e, V_e \in \mathbb{R}^{C{\times}C}$ are used to transform $F_s^{2\downarrow}$ into query $Q_e$, key $K_e$, and value $V_e \in \mathbb{R}^{\frac{{I_f}{J_f}}{4}{\times}C}$. $A_e$ is the attention output and $F_e$ denotes the encoder feed-forward network implemented with position-wise fully connective layers. Finally, we obtain a global flow-aware feature $F_l^e \in \mathbb{R}^{\frac{{I_f}{J_f}}{4}{\times}C}$, where $F_l^e[i]$ is the long-range feature of the $i$-th region and it contains the traffic flow information from other regions on the basis of flow similarity.

In the transformer decoder, we adopt the road network feature $F_r$ as a query priori to generate a global road-aware feature. Specifically, $F_r$ is also downsampled with a $2{\times}2$ max-pooling layer and reshaped to $\frac{{I_f}{J_f}}{4}{\times}C$. The processed road feature is first fed into a self-attention unit for the query priori embedding, whose output is denoted as $F_r^q \in \mathbb{R}^{\frac{{I_f}{J_f}}{4}{\times}C}$. We then perform the long-range road-aware inference with the following formulation:
\begin{equation}
\begin{aligned}
Q_d &= F_r^q*\mathcal{W}_d^q,{~~}K_d = F_l^e*\mathcal{W}_d^k,{~~}V_d = F_l^e*\mathcal{W}_d^q, \\
A_d &= F_r^q+softmax(\frac{Q_dK_d}{\sqrt{C}})V_d, \\
F_l &= A_d + F_d(A_d),
\end{aligned}
\end{equation}
where the decoder query $Q_d \in \mathbb{R}^{\frac{{I_f}{J_f}}{4}{\times}C}$ is generated from road network priori, the key $K_d$, and value $V_d \in \mathbb{R}^{\frac{{I_f}{J_f}}{4}{\times}C}$ are generated from the flow-aware feature $F_l^e$. $F_d$ is the decoder feed-forward network. $F_l \in \mathbb{R}^{\frac{{I_f}{J_f}}{4}{\times}C}$ is our long-range  road-aware feature. Notice that $F_l[i]$ is the long-range feature of the $i$-th region, which contains the traffic flow information from other regions on the basis of both road network distribution and flow similarity.

Then, $F_l$ is reshaped back to $\frac{I_f}{2}{\times}\frac{J_f}{2}{\times}C$ and resized back to the original resolution $I_f{\times}J_f$ with bilinear interpolation. To simultaneously incorporate short-range and long-range information, the feature $F_s$ and the processed $F_l$ are fused with an element-wise addition operator. Finally, the fused feature is fed into a $1{\times}1$ convolutional layer to estimate our fine-grained traffic flow map $\hat{Y}_t \in \mathbb{R}^{I_f{\times}J_f{\times}S}$ for the time interval $t$.

\section{Experiments}\label{sec:experiment}
In this section, we first introduce the settings of our experiments (e.g., dataset construction, implementation details, and evaluation metrics). We then compare the proposed RATFM with eight representative approaches under various scenarios, including the comparisons on the whole testing sets, in heavy traffic regions, and over different time periods. Finally, we perform an internal analysis to verify the effectiveness of each component of our method.

\subsection{Experiments Settings}
\subsubsection{Dataset Construction}
To the best of our knowledge, there are few public benchmarks (e.g., TaxiBJ \cite{liang2019urbanfm}) for fine-grained traffic flow inference. To promote the development of this field, we construct two large-scale datasets, i.e., XiAn and ChengDu, by collecting a mass of vehicle trajectories and external factor data from two real-world cities\footnote{https://gaia.didichuxing.com}. Notice that our datasets are more challenging because of the focus on inflow/outflow, where the fine-grained flow is much different from the coarse-grained flow and the total inflow/outflow of $N{\times}N$ fine-grained regions is not equal to the inflow/outflow of the corresponding coarse-grained region. By contrast, TaxiBJ focuses on the volume-based flow, where the distribution of fine-grained flow is similar to that of coarse-grained flow and the total volume of $N{\times}N$ fine-grained regions is equal to the volume of the corresponding coarse-grained region. In this work, we conduct experiments on all the above datasets, whose overviews are summarized in Table \ref{dataset}. The HappyValley dataset \cite{liang2019urbanfm} is not adopted in our work, since Happy Valley is a theme park without obvious traffic roads. Moreover, this dataset has not been released.

\begin{table}
  \caption{The overview of the XiAn, ChengDu, and TaxiBJ-P1 datasets for fine-grained traffic flow inference.}
  \vspace{-2mm}
  \label{dataset}
  \resizebox{\linewidth}{!} {
  \begin{tabular}{cccc}
    \toprule
    \textbf{Dataset}    & \textbf{XiAn}       & \textbf{ChengDu}       & \textbf{TaxiBJ-P1}  \\
    \midrule
    City                & Xian, China             & Chengdu, China      & Beijing, China       \\
    Time Interval       & 15min                   & 15min               & 30min                      \\
    Flow Type           & Inflow\&Outflow         & Inflow\&Outflow     & Volume                \\
    Coarse-grained Size & 16$\times$16$\times$2   & 16$\times$16$\times$2 & 32$\times$32$\times$1                   \\
    Fine-grained Size   & 64$\times$64$\times$2   & 64$\times$64$\times$2 & 128$\times$128$\times$1                  \\
    Year                & 2016                    &2016                 &2013                   \\
    Training Timespan   & 10/01-11/02             & 10/01-10/21         &07/01-08/31                   \\
    Validation Timespan & 11/03-11/16             & 10/22-10/24         &09/01-09/30                   \\
    Testing Timespan    & 11/17-11/30             & 10/25-10/31         &10/01-10/31          \\ \hline
    Rendered Road Network & 928$\times$928        & 952$\times$952      & 1125$\times$1125  \\
    Day of Week         & {[}0, 6{]}              & {[}0, 6{]}          & {[}0, 6{]}     \\
    Time of Hour        & {[}0, 23{]}             & {[}0, 23{]}         & {[}7, 21{]}     \\
    Weather             & 14 types                & 14 types            & 16 types    \\
    Temperature/$^{\circ}$C & {[}-5, 32.2{]}      &{[}2.2, 32.2{]}      & {[}-24.6, 41.0{]}    \\
    Windspeed/mph       & {[}0, 18{]}             & {[}0, 22{]}         & {[}0, 48.6{]}    \\
    \bottomrule
    \end{tabular}
  }
\end{table}

\begin{figure}[t]
  \centering
  \includegraphics[width=1\linewidth]{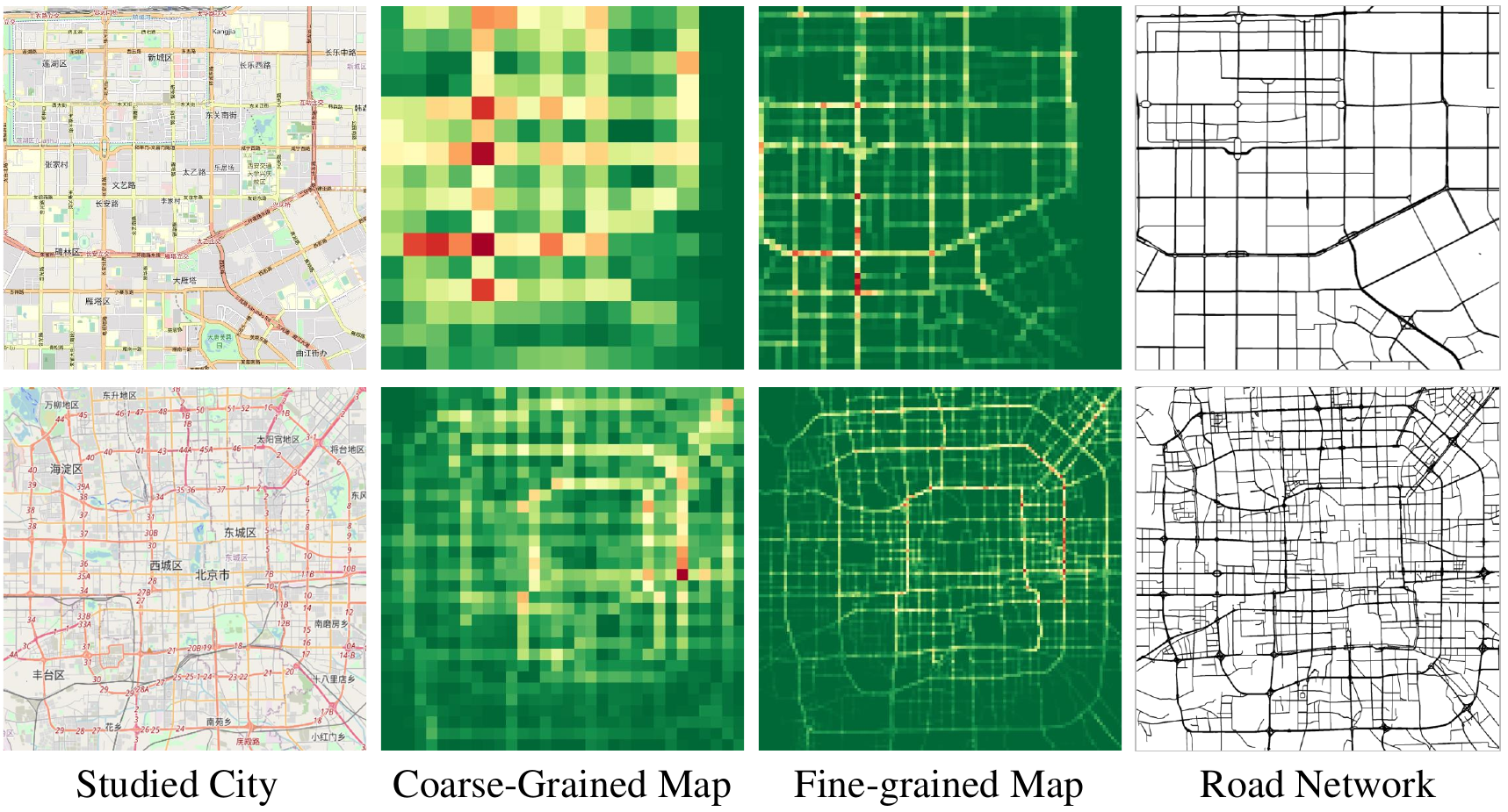}
  \vspace{-7mm}
  \caption{Visualization of the urban map, coarse-grained traffic flow map, fine-grained flow map, and road network map for the XiAn dataset (top row) and the TaxiBJ dataset (bottom row).}
  \label{fig:visual_datasets}
\end{figure}

\textbf{1) XiAn}: This dataset was built based on the vehicle flow of Xian, China. An area of 20 square kilometers within the second ring road is taken as our studied area. This area is split into $16\times16$ grids for the coarse-grained map, while the fine-grained map is composed of $64\times64$ grids. A total of 1221 million records were collected from Oct.1st, 2016 to Nov.30th, 2016. Each record contains the information of vehicle ID, geographic coordinate, and the corresponding timestamp. For each grid, we measured its inflow and outflow every 15 minutes by counting the number of vehicles entering or exiting the grid. The data of the first month and that of the last two weeks are used for training and testing respectively, while the data of the remaining days are used for validation. Some traffic flow maps and the road network of our XiAn dataset are shown in the first row of Fig. \ref{fig:visual_datasets}.

\textbf{2) ChengDu}:
This dataset was created based on the 32 million vehicle trajectory records of Chengdu, China, which were collected from Oct. 1st, 2016 to Oct. 31th 2016. The resolutions of coarse-grained and fine-grained maps are set to $16\times16$ and $64\times64$, respectively. Each time interval is 15 minutes. Moreover, this dataset is divided into a training set (10/01 - 10/21), a validation set (10/22 - 10/24), and a testing set (10/25 - 10/31). Some traffic flow maps and the road network of ChengDu are shown in Fig. \ref{fig:intro}.

\textbf{3) TaxiBJ}: This dataset was released by Liang et al. \cite{liang2019urbanfm} and it was constructed with a large number of taxi trajectories in Beijing, China. Specifically, the studied area was divided into $32\times32$ grids for coarse-grained maps and $128\times128$ grids for fine-grained maps. For each grid, the taxi volume was measured every 30 minutes, thus the channel number of its traffic flow maps is 1. The external factors of this dataset are processed by Liang et al. \cite{liang2019urbanfm} using their own strategies. According to the time of data collection, this dataset was divided into four parts, including P1 (7/1/2013-10/31/2013), P2 (2/1/2014-6/30/2014), P3 (3/1/2015-6/30/2015), and P4 (11/1/2015-3/31/2016). Moreover, each part was divided into a training set, a validation set, and a testing set by a ratio of 2:1:1. Some traffic flow maps and the road network of TaxiBJ are shown in the second row of Fig. \ref{fig:visual_datasets}. In the following sections, we would conduct extensive evaluations on TaxiBJ-P1, while conducting standard evaluations on other parts, since these parts have similar traffic flow distributions.

\subsubsection{Implementation Details}
In this work, the proposed RATFM is implemented with the popular deep learning framework PyTorch \cite{paszke2019pytorch}. Most of our hyper-parameter settings are the same as the default settings in UrbanFM \cite{liang2019urbanfm}. Specifically, the channel number $C$ is set to 128 uniformly, while the radius $R$ of 1D convolutional layers is set to 4. The filter weights of all layers are initialized by Xavier\cite{glorot2010understanding}. The learning rate is initialized to 2e-4 and the decay ratio is $0.5$. The batch size is set to 8 for XiAn, 4 for ChengDu, and 4 for TaxiBJ. Notice that, in the long-range road-aware inference module, we adopt $4{\times}4$ max-pooling layers for TaxiBJ to reduce the computational cost, since the resolution of TaxiBJ is much higher than that of the XiAn and ChengDu datasets. Adam\cite{kingma2014adam} is utilized to optimize our network by minimizing the Mean Square Error between the estimated fine-grained maps and the corresponding ground-truths. On each benchmark, we train the proposed RATFM with the training set, and use the validation set to decide which epoch model to save. Finally, the well-trained model is evaluated on the testing set.

\subsubsection{Evaluation Metrics}
In this work, we evaluate the performance of different methods using Root Mean Square Error (RMSE), Mean Absolute Error (MAE), and Mean Absolute Percentage Error (MAPE), which are defined as follows:
\begin{equation}
  RMSE = \sqrt{\frac{1}{n}\sum_{i=1}^{n}\left ( \hat{Y_{i}}- Y_{i}\right )^2}
\end{equation}
\begin{equation}
  MAE = \frac{1}{n}\sum_{i=1}^{n}\left \| \hat{Y_{i}}- Y_{i} \right \|
\end{equation}
\begin{equation}
  MAPE = \frac{1}{n}\sum_{i=1}^{n}\frac{ \sum \left \| \hat{Y_{i}}- Y_{i} \right \|}{\sum \left \| Y_{i}\right \|}
  \label{eq:metrics_MAPE}
\end{equation}
where $n$ is the total number of samples, $\hat{Y_{i}} $, and $Y_{i}$ denote the $i$-th estimated flow map and the corresponding ground-truth map, respectively. Notice that our MAPE is different from the position-wise MAPE adopted in \cite{liang2019urbanfm}, which is sensitive and unstable for those regions with low flow volume. To eliminate this issue, we use citywide MAPE, where the sum of flow errors of all regions is divided by the sum of the ground-truth flow of all regions.

\subsection{Comparison with State-of-the-Art Methods}
In this subsection, we compare the proposed RATFM with eleven previous models, including the statistics Historical Average (HA), six representative methods for image processing (e.g., VDSR \cite{kim2016accurate}, ESPCN \cite{shi2016real}, SRCNN \cite{dong2015image}, SRResNet \cite{ledig2017photo}, DeepSD \cite{vandal2017deepsd}, and IPT \cite{chen2021pre}), and four current methods for fine-grained traffic flow inference (e.g., UrbanFM \cite{liang2019urbanfm}, UrbanPy \cite{ouyang2020fine}, UrbanODE \cite{zhou2021inferring}, and CUFAR \cite{yu2023Overcoming}). The details of these methods are described in our supplementary material.

\begin{table}[t]
  \caption{Performance of different methods on the whole testing sets of XiAn and ChengDu.}
  \vspace{-2mm}
  \centering
  \resizebox{1\linewidth}{!} {
  \begin{tabular}{c|ccc|ccc}
  \hline
  \multirow{2}{*}{Method}           & \multicolumn{3}{c|}{\textbf{XiAn}}       & \multicolumn{3}{c}{\textbf{ChengDu}}   \\
  \cline{2-7}
              & RMSE & MAE & MAPE & RMSE & MAE & MAPE \\
  \hline
  HA                               & 54.909 & 21.897 & 318.50\% & 78.391 & 32.734 & 215.96\% \\
  VDSR \cite{kim2016accurate}      & 21.850 & 10.591 & 40.44\%  & 28.911 & 14.793 & 36.00\%  \\
  ESPCN \cite{shi2016real}         & 18.316 & 8.788  & 37.35\%  & 29.843 & 14.376 & 36.68\%  \\
  SRCNN \cite{dong2015image}       & 17.813 & 8.416  & 35.36\%  & 24.217 & 11.376 & 28.75\%  \\
  IPT  \cite{chen2021pre}          & 16.624 & 7.506  & 30.79\%  & 22.849 & 10.743 & 26.80\%  \\
  UrbanODE \cite{zhou2021inferring}& 19.371 & 7.886  & 28.30\%  & 25.972 & 11.064 & 24.98\%  \\
  SRResNet \cite{ledig2017photo}   & 16.946 & 6.988  & 27.76\%  & 21.809 & 9.690  & 23.65\%  \\
  DeepSD \cite{vandal2017deepsd}   & 16.617 & 6.566  & 26.09\%  & 21.653 & 9.096  & 22.12\%  \\
  UrbanFM \cite{liang2019urbanfm}  & 17.406 & 6.285  & 23.91\%  & 21.798 & 8.504  & 20.41\%  \\
  CUFAR \cite{yu2023Overcoming}    & 17.176 & 6.235  & 22.83\%  & 22.373 & 8.946  & 20.32\% \\
  UrbanPy \cite{ouyang2020fine}    & 16.521 & 5.933  & 22.03\%  & 21.817 & 8.473  & 19.49\% \\
  RATFM                            & {\textbf{16.260}} & {\textbf{5.746}}  & {\textbf{21.48\%}}  & {\textbf{21.248}} & {\textbf{8.033}}  & {\textbf{18.65\%}} \\
  \hline
  \end{tabular}
  \label{tab:whole_set}
  }
\end{table}

\begin{table}[t]
  \caption{Performance of different methods on the whole testing sets of TaxiBJ-P1. The results in brackets are quoted from previous works \cite{liang2019urbanfm,ouyang2020fine,zhou2021inferring,yu2023Overcoming}.}
  \vspace{-2mm}
  \centering
  %\resizebox{0.85\linewidth}{!} {
  \begin{tabular}{c|ccc}
  \hline
  \multirow{2}{*}{Method} &  \multicolumn{3}{c}{\textbf{TaxiBJ-P1}}   \\
  \cline{2-4}                      & RMSE & MAE & MAPE \\
  \hline
  HA                               & 9.998~~~~~~ & 4.204~~~~~~ & 39.88\% \\
  SRCNN \cite{dong2015image}       & 5.830 {\tiny(4.297)} & 3.651 {\tiny(2.491)} & 32.04\% \\
  ESPCN \cite{shi2016real}         & 4.187 {\tiny(4.206)} & 2.489 {\tiny(2.497)} & 21.60\% \\
  SRResNet \cite{ledig2017photo}   & 4.210 {\tiny(4.164)} & 2.510 {\tiny(2.457)} & 21.90\% \\
  DeepSD \cite{vandal2017deepsd}   & 4.094 {\tiny(4.156)} & 2.348 {\tiny(2.368)} & 20.39\% \\
  VDSR \cite{kim2016accurate}      & 4.265 {\tiny(4.159)} & 2.313 {\tiny(2.213)} & 20.02\% \\

  IPT \cite{chen2021pre}           & 4.111~~~~~~ & 2.372~~~~~~  & 20.40\% \\
  UrbanPy \cite{ouyang2020fine}    & 3.956 \tiny(3.949) & 1.994 \tiny(1.997) & 18.57\% \\
  UrbanODE \cite{zhou2021inferring}& 4.030 \tiny(3.852) & 2.112 \tiny(1.958) & 18.27\% \\
  UrbanFM \cite{liang2019urbanfm}  & 4.079 \tiny(3.950) & 2.100 \tiny(2.011) & 18.05\% \\
  CUFAR \cite{yu2023Overcoming}    & 3.880 \tiny(3.872) & 1.955 \tiny(1.952) & 16.84\% \\
  RATFM                            & {\textbf{3.846}}~~~~~~        & {\textbf{1.951}}~~~~~~        & {\textbf{16.83\%}}\\
  \hline
  \end{tabular}
  \label{tab:TaxiBJ-P1}
  %}
\end{table}

\begin{table}[t]
  \caption{Performance of different methods on the whole testing sets of TaxiBJ P2-P4. The results of those compared methods are directly quoted from \cite{liang2019urbanfm,ouyang2020fine,yu2023Overcoming}. Notice that MAPE isn't compared here, since its definition in our work is slightly different from that of other works.}
  \vspace{-2mm}
  \centering
  \resizebox{1\linewidth}{!} {
  \begin{tabular}{c|cc|cc|cc}
  \hline
  \multirow{2}{*}{Method}           & \multicolumn{2}{c|}{\textbf{TaxiBJ-P2}}       & \multicolumn{2}{c}{\textbf{TaxiBJ-P3}}  & \multicolumn{2}{c}{\textbf{TaxiBJ-P4}} \\
  \cline{2-7}
              & RMSE & MAE & RMSE & MAE & RMSE & MAE \\
  \hline
  VDSR \cite{kim2016accurate}      & 4.586 & 2.498 & 4.730 & 2.548 & 3.654 & 1.978  \\
  ESPCN \cite{shi2016real}         & 4.569 & 2.727 & 4.744 & 2.862 & 3.728 & 2.228  \\
  SRCNN \cite{dong2015image}       & 4.612 & 2.681 & 4.815 & 2.829 & 3.838 & 2.289  \\
  SRResNet \cite{ledig2017photo}   & 4.524 & 2.660 & 4.690 & 2.775 & 3.667 & 2.189  \\
  DeepSD \cite{vandal2017deepsd}   & 4.554 & 2.612 & 4.692 & 2.739 & 3.877 & 2.297  \\
  UrbanFM \cite{liang2019urbanfm}  & 4.329 & 2.224 & 4.496 & 2.318 & 3.501 & 1.815  \\
  UrbanPy \cite{ouyang2020fine}    & 4.359 & 2.227 & 4.519 & 2.319 & 3.514 & 1.821 \\
  CUFAR \cite{yu2023Overcoming}    & 4.273 & 2.186 & 4.394 & 2.243 & 3.417 & 1.758 \\
  \hline
  RATFM                            & 4.280 & 2.196 & 4.413 & 2.277 & 3.464 & 1.805\\
  \hline
  \end{tabular}
  \label{tab:TaxiBJ-P234}
  }
\end{table}

\begin{figure*}[ht]
  \centering
  \includegraphics[width=1\linewidth]{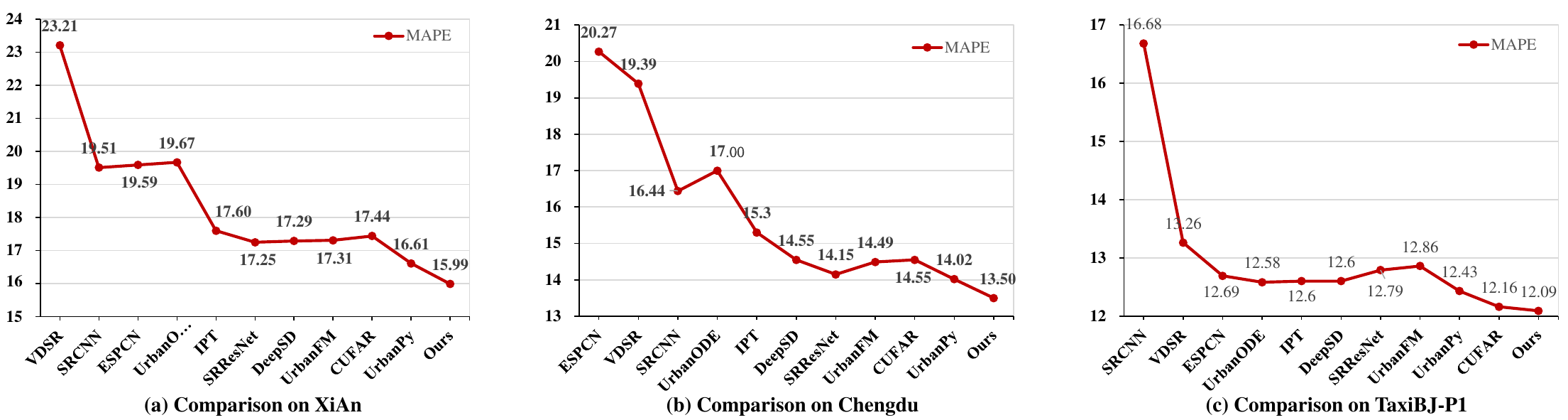}
  \vspace{-8mm}
  \caption{The MAPE of different deep learning-based methods in heavy traffic regions. Our RATFM achieves the state-of-the-art MAPE consistently on the XiAn, ChengDu and TaxiBJ-P1 datasets.}
  \label{fig:comparison_heavy_traffic}
\end{figure*}

\begin{figure*}[t]
  \centering
  \includegraphics[width=0.95\linewidth]{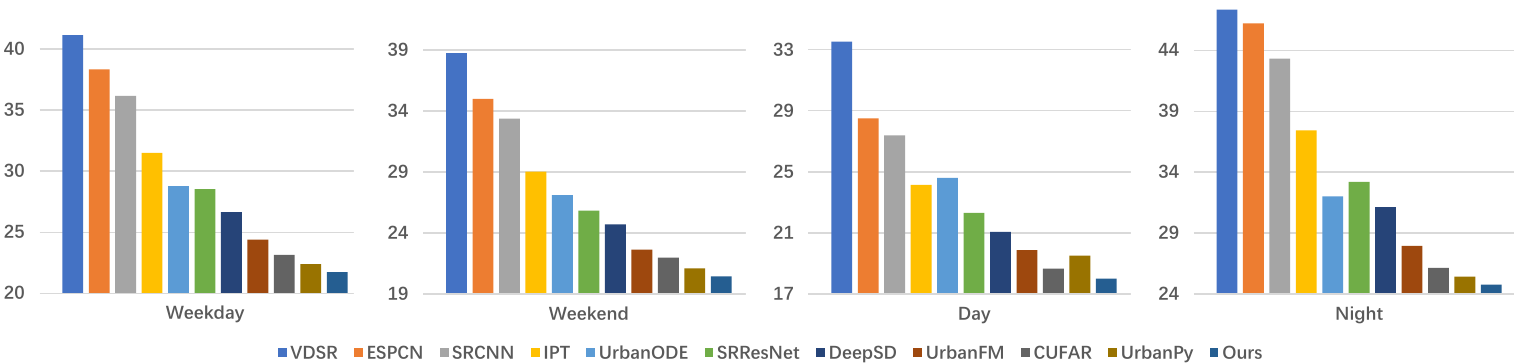}
  \vspace{-2mm}
  \caption{The MAPE of different methods for weekday, weekend, day and night on the XiAn dataset.}
  \label{fig:different_time_xian}
\end{figure*}

\begin{figure*}[t]
  \centering
  \includegraphics[width=0.95\linewidth]{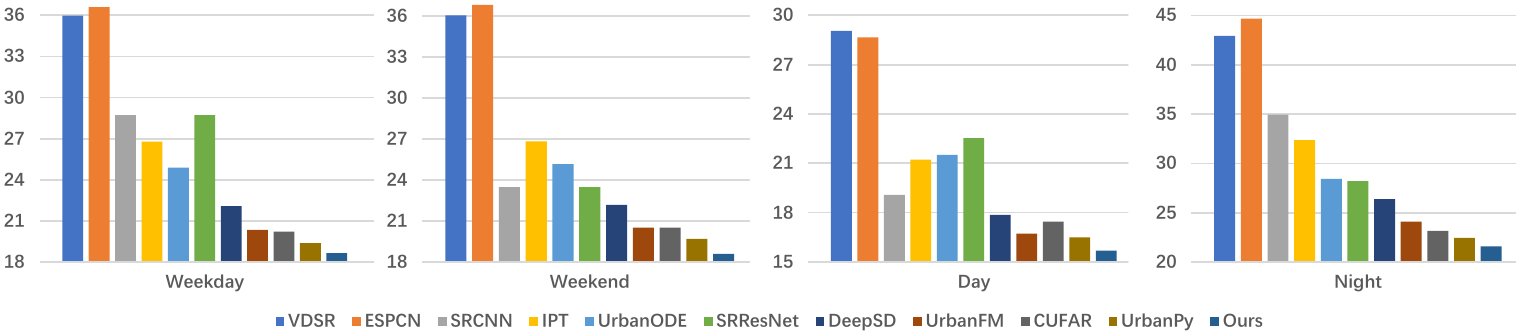}
  \vspace{-2mm}
  \caption{The MAPE of different methods for weekday, weekend, day and night on the ChengDu dataset.}
  \label{fig:different_time_chengdu}
\end{figure*}

Notice that some previous methods \cite{liang2019urbanfm,ouyang2020fine,zhou2021inferring,yu2023Overcoming} contain an optional $N^2$-Normalization operation. Specifically, these methods usually have three steps: 1) generate the coarse-to-fine mapping weights for each fine-grained region; 2) normalize the coarse-to-fine mapping weights among $N{\times}N$ fine-grained region; 3) apply the element-wise multiplication of coarse-grained traffic flow maps and the normalized mapping weights to obtain fine-grained traffic flow maps. However, the $N^2$ structural constraint only holds on the volume-based datasets, but does not hold on inflow/outflow-based datasets. Therefore, we reimplement these above methods using all steps for TaxiBJ, while using steps 1 and 3 for the XiAn and ChengDu datasets. By contrast, our method doesn't use the $N^2$-Normalization technique at all, i.e., directly using the long/short-range features to generate the fine-grained traffic flow maps. Moreover, those super-resolution based methods \cite{dong2015image,shi2016real,kim2016accurate,ledig2017photo,vandal2017deepsd,chen2021pre} also don't use the $N^2$-Normalization operation, i.e., they directly output the fine-grained maps.

\subsubsection{\bf{Comparison on the Whole Testing Sets}}
In this section, we compare the performances of all methods on the whole testing sets, i.e., all regions and all time periods. The results on the XiAn and ChengDu datasets are summarized in Table \ref{tab:whole_set}. We can observe that the baseline HA obtains unacceptable performance, especially on our challenging XiAn and ChengDu dataset, because this model ignores the dynamics of traffic flow completely. Compared with HA, those deep learning-based approaches perform better due to the superiority of deep representation learning. For instance, SRCNN obtains a MAPE 35.36\% on XiAn, while ESPCN gets a MAPE 36.68\% on ChengDu. The MAPE of the transformer-based IPT is 30.79\% on XiAn and 26.80\% on ChengDu. Despite progress, these common super-resolution methods still cannot achieve satisfactory results. With some architecture designed for fine-grained traffic flow modeling, some recent models (e.g. UrbanFM, CUFAR, and UrbanPy) can achieve competitive performance. For example, UrbanPy obtains a MAPE of 22.03\% on XiAn and 19.49\% on ChengDu. However, these methods simply use the data-driven strategy to roughly learn a mapping from coarse-grained data to fine-grained data, and this task still has a lot of room for improvement. To fully capture the traffic flow distribution patterns, our RATFM incorporates the road network as prior knowledge to learn semantic representation effectively, thereby achieving state-of-the-art performance on all datasets. For instance, our RATFM obtains the best MAPE 21.48\% on the XiAn dataset, and decreases the MAPE to 18.65\% on the ChengDu dataset.

The performance of different methods on TaxiBJ is summarized in Table \ref{tab:TaxiBJ-P1} and Table \ref{tab:TaxiBJ-P234}. It is worth noting that we reimplement all compared methods on TaxiBJ-P1 for extensive evaluations. On the other parts of the TaxiBJ dataset, the results of compared methods are directly quoted from \cite{liang2019urbanfm,ouyang2020fine,yu2023Overcoming} for standard evaluations. It can be observed that our method consistently outperforms all previous methods on all metrics on TaxiBJ-P1, and it can also achieve highly competitive results on the TaxiBJ P2-P4 datasets. These comparisons well demonstrate the effectiveness of the proposed method for fine-grained traffic flow inference.

\subsubsection{\bf{Comparison on Heavy Traffic Regions}}
As mentioned above, the distribution of urban traffic flow is uneven in space. In general, most of the traffic flow is distributed in road areas. Therefore, in this section, we focus on the fine-grained traffic flow inference in heavy traffic regions, because the traffic states of these regions are crucial for urban transportation management. Specifically, we first conduct a statistical analysis on fine-grained traffic flow maps of the training set and select the top $20\%$ grids with the highest flow as target regions. We then perform in-depth evaluations in these selected regions.

Fig. \ref{fig:comparison_heavy_traffic} summarizes the performances of all deep learning-based methods in heavy traffic regions. We can observe that the proposed RATFM can achieve the state-of-the-art MAPE consistently on the XiAn, ChengDu, and TaxiBJ-P1 datasets. Specifically, our RATFM obtains a MAPE 15.99\% in heavy traffic regions of XiAn and has a relative improvement of 3.7\%, compared with the competitor UrbanPy. On the ChengDu dataset, our RATFM obtains a MAPE 13.50\%, while the MAPE of UrbanPy is 14.02\%. On the TaxiBJ-P1 dataset, our method also outperforms other models with an impressive MAPE of 12.09\%. Such impressive performance is attributed  to that we introduce the road network prior and guide our network to better infer the fine-grained traffic distribution. The comparisons well demonstrate the stability of the proposed method in heavy traffic regions.

\subsubsection{\bf{Comparison on Different Time Periods}}
To verify the robustness of our method, we further compare the proposed RATFM with other approaches over different time periods. Notice that the detailed time information of TaxiBJ-P1 is not provided, thus we mainly conduct experiments on the ChengDu and XiAn datasets in this section.

\textbf{1) Weekday and Weekend: }
Following previous works \cite{liu2019contextualized,liu2020dynamic}, we first compare the inference results on weekdays and weekends, respectively, since the traffic patterns of weekdays are different from that of weekends to some extent. For instance, commuters (e.g, office workers and students) are the main force of traffic travels on weekdays and they tend to go to fixed places, while people usually go to various places on weekends. In this work, the weekday performance is represented with the mean MAPE from Monday to Friday, while the weekend performance is the mean MAPE of Saturday and Sunday. The performances of different approaches on weekday and weekend are summarized in Fig. \ref{fig:different_time_xian} and \ref{fig:different_time_chengdu}. We can observe that the proposed RATFM outperforms all comparison methods on both the XiAn and ChengDu datasets. Specifically, on XiAn, our RATFM obtains a MAPE 21.74\% for weekdays and 20.46\% for weekends, while the best MAPE of previous methods is 22.40\% for weekdays and 21.10\% for weekends. Moreover, our RATFM achieves a MAPE 18.66\% for weekdays and 18.61\% for weekends on ChengDu. Compared with the typical UrbanFM, our method has relative improvements of 8.3\% and 9.4\% for the corresponding time periods respectively. These comparisons show that our method can well capture the fine-grained traffic flow distribution regardless of traffic patterns.

\textbf{2) Day and Night: }
We then evaluate the fine-grained traffic flow inference during the days and nights respectively, since travel patterns may be much different at these two periods. In our work, the daytime is defined as 06:00-18:00, while the nighttime is defined as 18:00-06:00 till the next day. In Fig. \ref{fig:different_time_xian} and \ref{fig:different_time_chengdu}, we show the performance (MAPE) of different methods for both the days and nights. Specifically, on the XiAn dataset, our RATFM obtains a MAPE 18.00\% for days and 24.75\% for nights, while that of the previous best-performing method UrbanPy are 19.51\% and 25.41\%, respectively. On the ChendDu dataset, our RATFM can further decrease the MAPE from 16.50\% to 15.69\% for day and from 22.47\% to 21.60\% for nights. One interesting phenomenon is that the day MAPE is much lower than the night MAPE. This is because that most of the day travelers are office workers and students, whose mobility patterns are more regular. In general, whether for day or night, our method outperforms all comparison methods on both the XiAn and ChengDu datasets.

\begin{figure}[t]
  \centering
  \includegraphics[width=\linewidth]{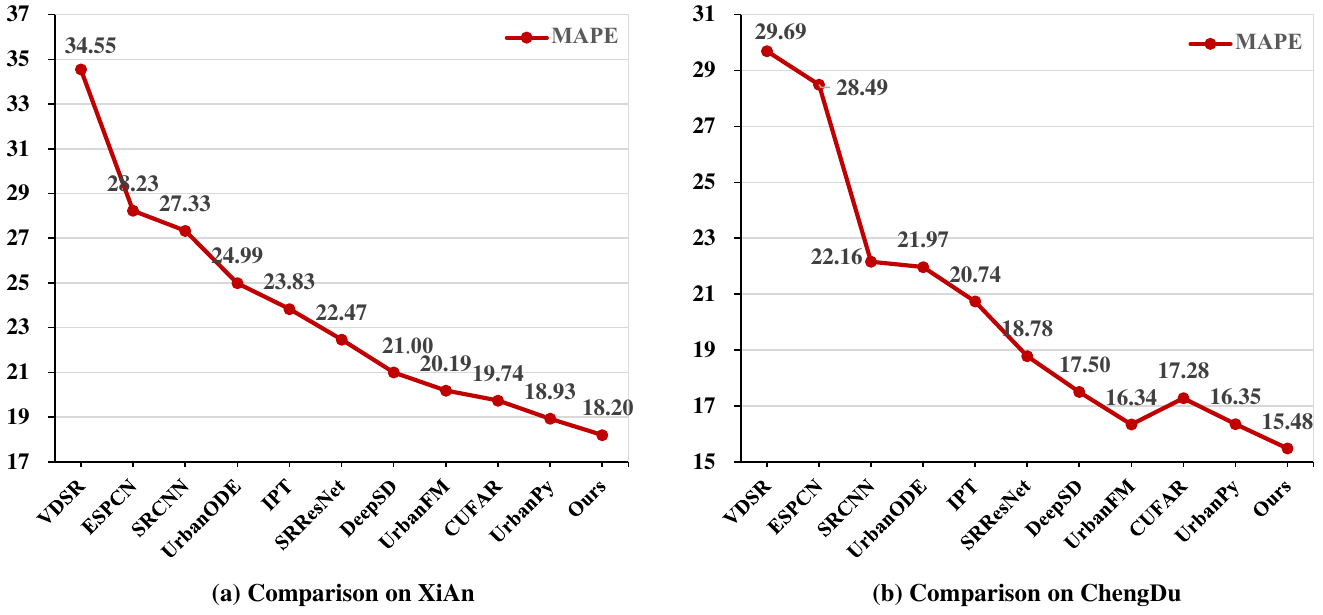}
  \vspace{-7mm}
  \caption{The MAPE of different methods during rush hours on XiAn and ChengDu datasets.}
  \label{rushtime}
\end{figure}

\begin{table*}[t]
  \caption{The performance of different variants of our RATFM on the XiAn, ChengDu, and TaxiBJ-P1 Datasets.}
  \vspace{-2mm}
  \label{tab:variants}
  \centering
  \resizebox{0.87\linewidth}{!} {
  \begin{tabular}{l|ccc|ccc|ccc}
  \hline
  \multirow{2}{*}{Model}              & \multicolumn{3}{c|}{XiAn}           & \multicolumn{3}{c|}{ChengDu}         & \multicolumn{3}{c}{TaxiBJ-P1}      \\
  \cline{2-10}
    & RMSE & MAE & MAPE & RMSE & MAE & MAPE & RMSE & MAE & MAPE \\
  \hline
   Short-Net                     & 17.255   & 6.252   & 23.94\%   & 23.014   & 8.717   & 20.44\%   & 4.077   & 2.081   & 18.00\%       \\
   Road+Short-Net                & 16.263   & 5.905   & 22.65\%   & 21.347   & 8.293   & 19.41\%   & 4.002   & 2.032   & 17.50\%       \\
   Road+Short+Long-Net           & 16.240   & 5.777   & 21.72\%   & 21.309   & 8.077   & 18.71\%   & 3.891   & 1.971   & 16.98\%       \\
   Road+Short+Long+External-Net  & \bf{16.181}   & \bf{5.731}   & \bf{21.38\%}   & \bf{21.248}   & \bf{8.034}   & \bf{18.65\%}   & \bf{3.846}   & \bf{1.951}   & \bf{16.83\%}      \\
   \hline
   ConcatRoad+Short+Long+External-Net & 17.843 & 6.082 & 22.13\% & 23.693 & 8.504 & 18.96\% & 4.041 & 1.976 & 17.03\% \\
  \hline
  \end{tabular}
  }
\end{table*}

\textbf{3) Rush Hours: }
Finally, we focus on the urban traffic flow inference during rush hours, which are defined as 7:30-9:30 and 17:30-19:30 in this work.
The performance of ten deep learning-based methods is presented in Fig. \ref{rushtime}. We can observe that the proposed RATFM outperforms all comparison methods on both XiAn and ChengDu. Specifically, on the XiAn dataset, our method obtains a MAPE 18.20\% for rush hours, while the MAPE of UrbanFM and UrbanPy are 20.19\% and 18.93\%, respectively.
There exists a similar situation of the performance comparison on the ChengDu dataset. With a MAPE 15.48\%, our RATFM also performs better than UrbanFM and UrbanPy, whose MAPE are 16.34\% and 16.35\%, respectively. These experiments well exhibit the robustness of our method during rush hours.

In summary, our method is capable to achieve state-of-the-art performance under various settings of time periods. Such performance is attributed to that our RATFM can effectively capture the inherent distribution of urban traffic flow with the prior knowledge of road networks regardless of the time evolution. These experiments well demonstrate the robustness of our method for fine-grained traffic flow inference.

\subsection{Component Analysis}

As mentioned in Section \ref{sec:method}, our RATFM is composed of a road network modeling module, a short-range road-aware inference module, a long-range road-aware inference module, and an external factor modeling module. To verify the effectiveness of each component, we implement five variants of RATFM, which are described as follows:
\begin{itemize}
  \item {\textbf{Short-Net}}: This variant is a baseline that only utilizes a simple short-range inference module to generate fine-grained traffic flow maps. Notice that the short-range inference module is road-independent, since the road network is not involved in this variant.
  \item {\textbf{Road+Short-Net}}: This variant learns the local spatial distribution of traffic flow under the guidance of road network information. Specifically, this network consists of the road network modeling module and short-range road-aware inference module.
  \item {\textbf{Road+Short+Long-Net}}: This variant utilizes the road network information to facilitate the short-range inference and long-range inference simultaneously. Specifically, this network is composed of the above modules other than the external factor modeling module.
  \item {\textbf{Road+Short+Long+External-Net}}: This variant is the full model of the proposed RATFM. Compared with other variants, this network further incorporates external factors to generate the fine-grained traffic flow maps.
  \item {\textbf{ConcatRoad+Short+Long+External-Net}}: This variant doesn't have the road network modeling module. Instead, the road network map is directly sized to $I_f{\times}J_f$ and then concatenated with the upsampled coarse-grained traffic flow map and the external factor features as input to the short/long-range modules.
\end{itemize}

We evaluate the performance of these variants on three datasets and summarize the results in Table \ref{tab:variants}. Only using the local information of coarse-grained maps,
the baseline Short-Net obtains a MAPE 23.94\% on XiAn, 20.44\% on ChengDu, and 18.00\% on TaxiBJ-P1, ranking last among all variants. We observe that the performance of Short-Net is very close to that of the model UrbanFM, since they adopt similar network architectures, i.e., stacked residual blocks. When incorporating the road network as prior knowledge, the variant Short+Road-Net obtains obvious performance improvements on all evaluation metrics. For example, the MAPE is decreased to 22.65\% on XiAn, 19.41\% on ChengDu, and 17.50\% on TaxiBJ-P1. These improvements are mainly attributed to the fact that the road network map provides a wealth of information about the spatial distribution of traffic flows. Further, the variant Road-Short+Long-Net achieves very competitive MAPE, i.e., 21.72\% on XiAn, 18.71\% on ChengDu, and 16.98\% on TaxiBJ-P1, when introducing the road network into transformer for global inference. As verified in section \ref{sec:transformer_expe}, our road-aware global inference is much better than the original global inference \cite{carion2020end}. Thus, we can conclude that performing short-range inference and long-range inference with road network prior is more effective for fine-grained traffic flow inference. Finally, when incorporating external factors, the variant Road+Short+Long+External-Net can achieve the best results on all datasets. For example, the MAPE is reduced to 21.38\% on XiAn, 18.65\% on ChengDu, and 16.83\% on TaxiBJ-P1. Thus we can conclude that the external factor modeling is meaningful for fine-grained traffic flow inference, although the improvements are relatively slight \cite{liang2019urbanfm}.

Finally, we verify the necessity of the road network modeling module by comparing the performance of the fourth and fifth variants in Table \ref{tab:variants}. We observed that the variant ConcatRoad+Short+Long+External-Net, which removes the road network modeling module and directly concatenates the road network map as input, exhibits performance degradation on all metrics when compared to our full model. This highlights the importance of utilizing a specialized module to capture the prior knowledge of the road network.

\begin{comment}
\begin{table}
  \centering
  \caption{Explore the effectiveness of one-dimensional convolution in the road network branch on XiAn, ChengDu and TaxiBJ-P1 Dataset}
  \label{tab:1d2d_3datasets}
  \resizebox{\linewidth}{!} {
    \begin{tabular}{ccccc}
      \toprule
      \textbf{}        & \textbf{Metric} & \textbf{RMSE} & \textbf{MAE} & \textbf{MAPE}                 \\
      \midrule
      \textbf{Xian}    & InN + RN\_1d    & 16.2631       & 5.9052       & 0.2265                        \\
      \textbf{}        & InN + RN\_2d    & 16.2777       & 6.0851       & 0.2318 \\ \hline
      \textbf{Chengdu} & InN + RN\_1d    & 21.3473       & 8.293        & 0.1941                        \\
      \textbf{}        & InN + RN\_2d    & 21.6127       & 8.3012       & 0.1973  \\ \hline
      \textbf{TaxiBJ-P1} & InN + RN\_1d    & 4.0024      & 2.0321       & 0.175                         \\
                       & InN + RN\_2d    & 4.0581        & 2.0725       & 0.1784    \\
      \bottomrule
      \end{tabular}
  }
\end{table}
\end{comment}

\begin{figure}[t]
  \centering
  \includegraphics[width=0.9\linewidth]{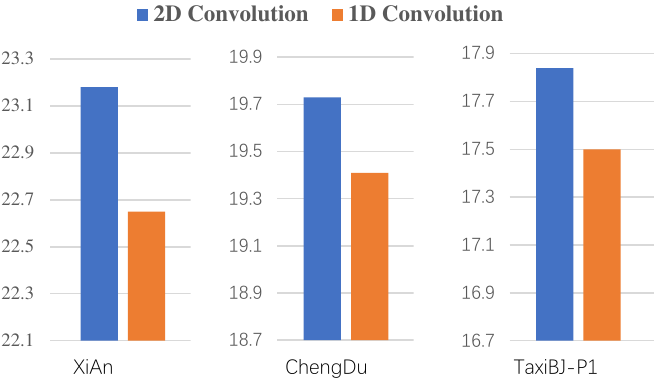}
  \vspace{-4mm}
  \caption{The MAPE of 1D convolution layers and 2D convolution layers for traffic road modeling.}
  \label{1D_vs_2D}
\end{figure}

\subsection{More Discussion}
\subsubsection{\bf{Effectiveness of 1D Convolution for Road Modeling}}
In Section \ref{sec:road_modeling}, we propose some special 1D convolutional layers to learn the semantic features of thin and long traffic roads on the map.
To verify the effectiveness of 1D convolution, we implement a variant of Short+Road-Net, which adopts the 2D convolutional layers with square filters in the road network branch. The performances of different convolutions are summarized in Fig. \ref{1D_vs_2D}. It can be observed that there is a certain performance degradation on each dataset when 2D convolution is used. For instance, the MAPE is increased from 22.65\% to 23.18\% on XiAn, from 19.41\% 19.73\%  on ChengDu, and from 17.50\% to 17.84\% on TaxiBJ-P1. This is due to that 2D convolutional layers involve too much information of background, while our 1D convolutional layers have some thin/long filters which are more aligned with road shapes. These experiments well demonstrate the effectiveness of our 1D convolutional layers for traffic road modeling.

\begin{figure}[t]
  \centering
  \includegraphics[width=0.875\linewidth]{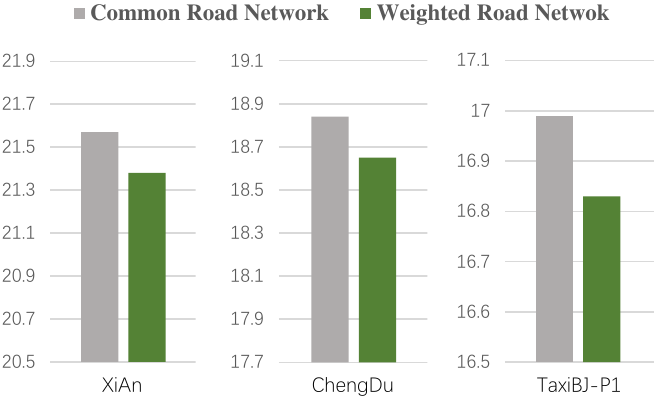}
  \vspace{-3mm}
  \caption{The MAPE of the weighted road network map and common road network map.}
  \label{Weighted_Road_Network}
\end{figure}

\subsubsection{\bf{Influences of the Weighted Road Network Map}}
In section \ref{sec:road_network_generation}, we notice that the traffic flow of each road is different, thus we generate a weighted road network on the basis of historical flow. Here we explore the influences of the weighted road network map and common road network map. Specifically, we implement two RATFM, one of which takes the weighted map as input and another one uses the common map. The performances of different road network maps are summarized in Fig. \ref{Weighted_Road_Network}. We can observe that the performance of the weighted map is better than that of the common map on all datasets. Thus we conclude that the weighted road network map is useful for fine-grained traffic flow inference to some extent.

\begin{figure}[t]
  \centering
  \includegraphics[width=0.85\linewidth]{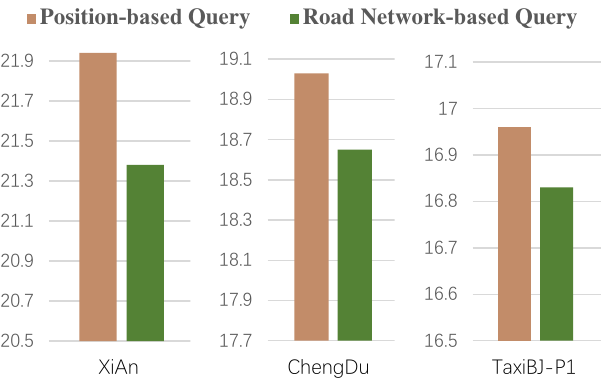}
  \vspace{-3mm}
  \caption{The MAPE of road network-based query and position-based query in the transformer decoder.}
  \label{Road_Network_Query}
\end{figure}

\subsubsection{\bf{Effectiveness of the Road Network-based Query}}\label{sec:transformer_expe}
In Section \ref{sec:long_range}, we take the road network feature as query priori, which is fed into the transformer decoder for global feature generation. In the previous work \cite{carion2020end}, a learned position feature is adopted as input of the transformer decoder. Here we compare the performance of road network-based query and position-based query. As shown in Fig. \ref{Road_Network_Query}, our RATFM achieves better performance on all datasets when using the road network information. For instance, our road network-based query obtains a MAPE 21.38\% on the XiAn dataset, while the position-based query has a MAPE 21.94\%. On the ChengDu dataset, with a MAPE 18.65\%, our road network-based query also outperforms the position-based query, whose MAPE is 19.03\%. In summary, the road network-based query can improve the performance with some considerable margins.

\begin{table}[t]
  \caption{The MAPE of different numbers of transformer layers.}
  \vspace{-2mm}
  \centering
  \begin{tabular}{c|ccc}
  \hline
  Layer Number &  XiAn & ChengDu & TaxiBJ-P1 \\
  \hline
  1     & 21.75\% & 18.84\%      & 16.92\% \\
  2     & 21.48\% & \bf{18.65\%} & \bf{16.83\%} \\
  3     & 21.62\% & 18.80\%      & 16.90\% \\
  \hline
  \end{tabular}
  \label{transformer_layer}
  %}
\end{table}

\begin{table}[t]
  \caption{The MAPE of different numbers of transformer heads.}
  \vspace{-2mm}
  \centering
  \begin{tabular}{c|ccc}
  \hline
  Head Number &  XiAn & ChengDu & TaxiBJ-P1 \\
  \hline
  1     & 21.93\% &  18.93\%  &  16.93\%  \\
  2     & 21.71\% &  18.87\%  &  16.93\% \\
  4     & 21.68\% &  18.78\%  &  16.93\%  \\
  8     & \bf{21.48\%} & \bf{18.65\%} & \bf{16.83\%} \\
  16    & 21.79\% &  18.94\%  &  16.92\% \\
  \hline
  \end{tabular}
  \label{transformer_head}
  %}
\end{table}

\subsubsection{\bf{The Configuration of Transformer}}
We further explore the transformer configuration (e.g., layer number and head number) in the long-range road-aware inference module. As shown in Table \ref{transformer_layer}, our method can obtain competitive results when using a single-layer transformer, and achieves the best performance with a two-layer transformer. Regarding the head setting, our performance improves gradually as the head number increases, as shown in Table \ref{transformer_head}. The proposed method achieves the best performance when the head number is set to 8. Therefore, the transformer layer number and head number are set to 2 and 8, respectively, in our work.

\subsubsection{\bf{Efficiency Analysis}}
Here we compare the inference efficiency of different methods for fine-grained traffic flow inference. Specifically, we evaluate the inference time of six deep learning-based methods on the same NVIDIA 3090 GPU. As shown in Table \ref{efficiency}, CUFAR only costs 4.65$\sim$5.25 milliseconds for each inference, ranking first among all methods. Our RATFM requires 16.73 milliseconds on XiAn\&ChengDu and 24.94 milliseconds on TaxiBJ-P1 for each inference, which is more efficient than IPT and UrbanODE. In summary, all methods can run in real time and the inference efficiency is not the bottleneck of this task. We should focus on how to improve the performance of fine-grained traffic flow inference.

\begin{table}[t]
  \caption{The inference time (millisecond) of different methods for fine-grained traffic flow inference.}
  \vspace{-2mm}
  \centering
  %\resizebox{0.85\linewidth}{!} {
  \begin{tabular}{c|cc}
  \hline
  Method &  XiAn\&ChengDu & TaxiBJ-P1 \\
  \hline
  CUFAR \cite{yu2023Overcoming}     & ~~4.65& ~~5.25 \\
  UrbanFM \cite{liang2019urbanfm}   & 10.88 & 14.51 \\
  UrbanPy \cite{ouyang2020fine}     & 16.72 & 17.35 \\
  IPT \cite{chen2021pre}            & 27.42 & 30.61 \\
  UrbanODE \cite{zhou2021inferring} & 28.53 & 37.18 \\
  \hline
  RATFM                             & 16.73 & 24.94 \\
  \hline
  \end{tabular}
  \label{efficiency}
  %}
\end{table}

\subsubsection{\bf{Fine-grained Traffic Flow Prediction}}
Finally, we apply the proposed method to address fine-grained traffic flow prediction. Formally, the coarse-grained maps $\{Y_{t-P+1}, ..., Y_{t}\}$ of previous $P$ time and the current external factor $E_t$ are used to forecast the future fine-grained map $Y_{t+1}$, where the hyper-parameter $P$ is set to 4 in our work. Here we reimplement our RATFM and four recent deep models \cite{liang2019urbanfm,zhou2021inferring,chen2021pre,yu2023Overcoming} on the XiAn and ChengDu datasets. We don't use TaxiBJ-P1, since the time intervals of its samples are not continuous. More specifically, all coarse-grained maps are first concatenated on the channel dimension and then fed into the networks of different methods. Notice that the coarse-grained map $X_{t+1}$ is unavailable and it is meaningless for those methods \cite{liang2019urbanfm,zhou2021inferring,yu2023Overcoming} to generate coarse-to-fine mapping weights. To solve this issue, these methods are reimplemented to directly forecast the future map $Y_{t+1}$ as our method does. As shown in Table \ref{prediction}, our method achieves a MAPE 32.99\% on XiAn and 27.84\% on ChengDu, outperforming other methods by large margins. This experiment well demonstrates the effectiveness of the proposed method for fine-grained traffic flow prediction.

\begin{table}[h]
  \caption{The MAPE of different methods for fine-grained traffic flow prediction.}
  \vspace{-2mm}
  \centering
  %\resizebox{0.85\linewidth}{!} {
  \begin{tabular}{c|cc}
  \hline
  Method &  XiAn & ChengDu \\
  \hline
  IPT \cite{chen2021pre}            & 41.90\% & 34.81\% \\
  UrbanODE \cite{zhou2021inferring} & 41.53\% & 34.88\% \\
  UrbanFM \cite{liang2019urbanfm}   & 36.25\% & 31.06\% \\
  CUFAR \cite{yu2023Overcoming}     & 33.98\% & 29.66\% \\
  \hline
  RATFM                             & {\bf{32.99\%}} & {\bf{27.84\%}} \\
  \hline
  \end{tabular}
  \label{prediction}
  %}
\end{table}

\section{Conclusion}\label{sec:conclusion}
In this work, we focus on the task of fine-grained traffic flow inference, which aims to generate fine-grained data from coarse-grained data collected through limited traffic sensors. We observe that the traffic flow distribution is significantly similar to the road network structure, which can be considered as valuable prior knowledge for traffic flow inference. To this end, we propose a novel Road-Aware Traffic Flow Magnifier (RATFM) to estimate fine-grained flow under the gaudiness of road network information.
We construct a high-resolution road network map based on the crawled realistic geographic map and extract the road network feature with a tailor-designed multi-directional 1D convolutional layer. The extracted road feature is then incorporated into a stacked residual branch and a transformer architecture to fully learn the short/long-range spatial distribution of fine-grained traffic flow. The extensive experiments conducted on three real-world benchmarks show that the proposed RATFM is capable to achieve state-of-the-art performance under various scenarios.

\section*{Acknowledgement}
We would like to thank Didi Chuxing for providing the trajectory data of ChengCu and XiAn, China.

\ifCLASSOPTIONcompsoc
  %\section*{Acknowledgments}
\else
  %\section*{Acknowledgment}
\fi

\ifCLASSOPTIONcaptionsoff
  \newpage
\fi

\bibliographystyle{IEEEtran}
\bibliography{Traffic-Map-Reference}

\begin{IEEEbiography}[{\includegraphics[width=1in,height=1.25in,clip,keepaspectratio]{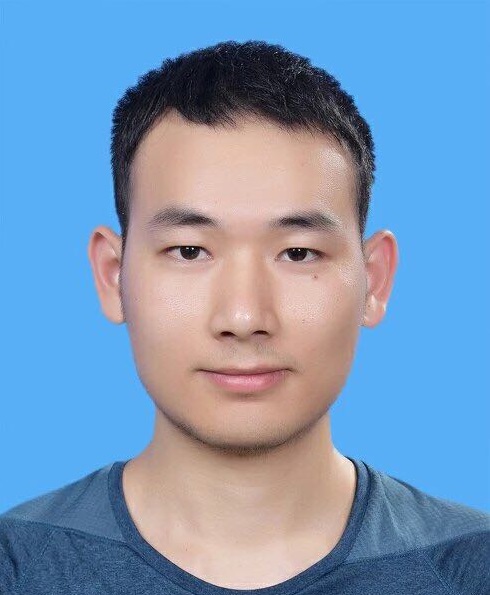}}]{Lingbo Liu}
received the Ph.D. degree from the School of Computer Science and Engineering, Sun Yat-sen University, Guangzhou, China, in 2020. From March 2018 to May 2019, he was a research assistant at the University of Sydney, Australia. He is currently a research assistant professor at Department of Land Surveying and Geo-Informatics, the Hong Kong Polytechnic University. His research interests include machine learning and urban computing. He has authorized and co-authorized on more than 20 papers in top-tier academic journals and conferences. He has been serving as a reviewer for numerous academic journals and conferences such as TPAMI, TKDE, TNNLS, TITS, CVPR, ICCV and IJCAJ.
\end{IEEEbiography}

\begin{IEEEbiography}[{\includegraphics[width=1in,height=1.25in,clip,keepaspectratio]{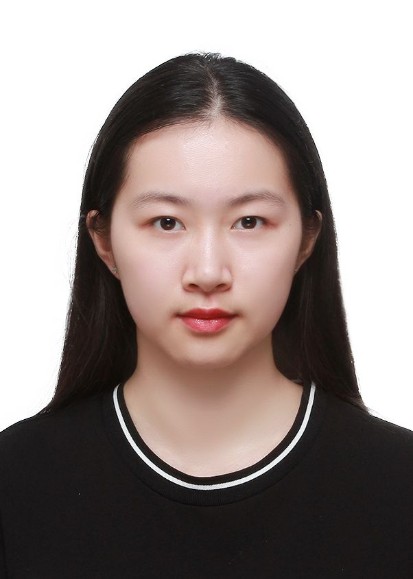}}]
{Mengmeng Liu}
received the B.E. degree from the School of Computer Science and Engineering, Sun Yat-sen University, Guangzhou, China, in 2019, where she is currently pursuing the Master's degree in Engineering (Computer Technology). Her current research interests include intelligent transportation systems and data mining.
\end{IEEEbiography}

\begin{IEEEbiography}[{\includegraphics[width=1in,height=1.25in,clip,keepaspectratio]{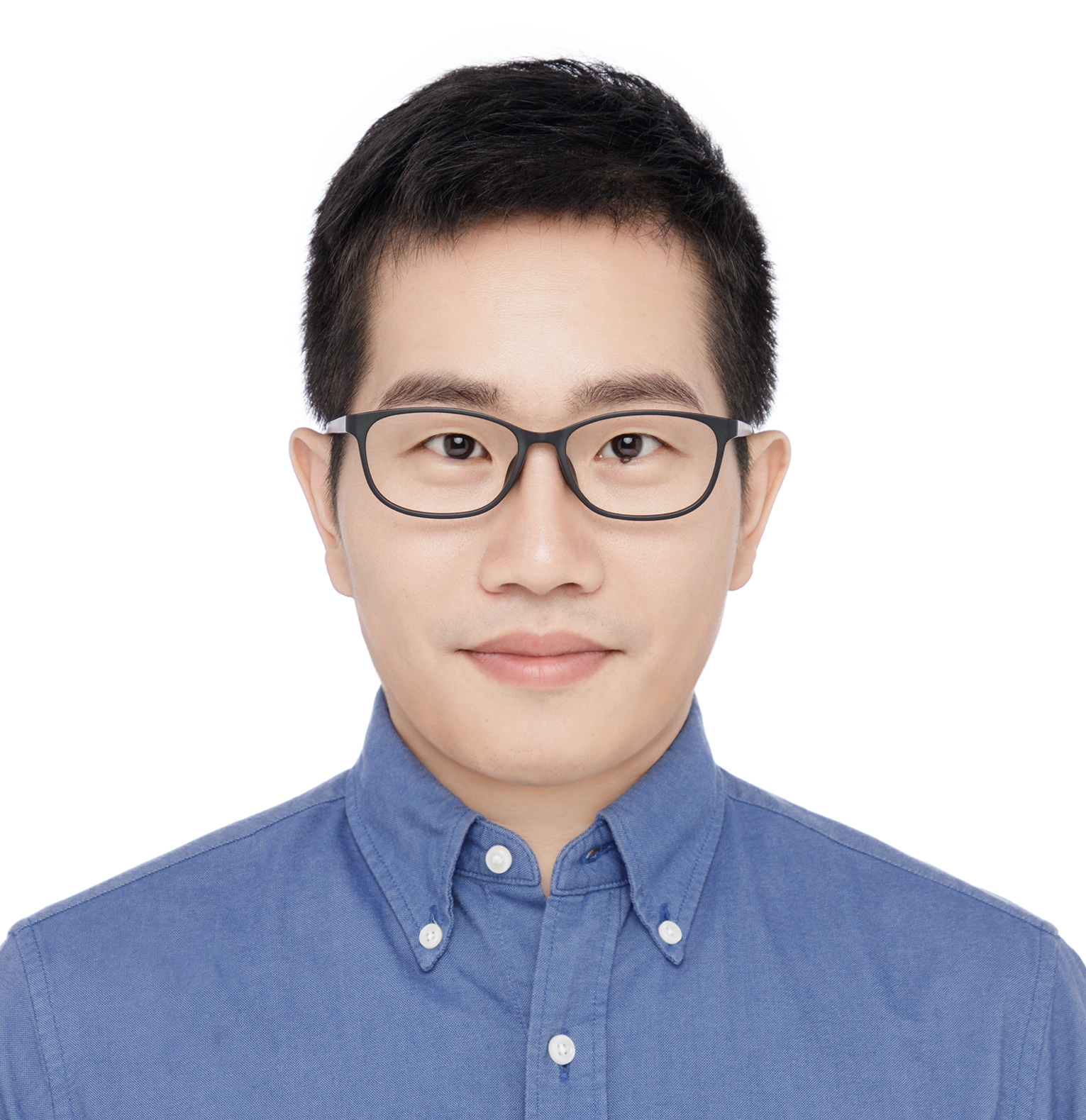}}]{Guanbin Li} (M'15) is currently an associate professor in School of Data and Computer Science, Sun Yat-sen University. He received his PhD degree from the University of Hong Kong in 2016. His current research interests include computer vision, image processing, and deep learning. He is a recipient of ICCV 2019 Best Paper Nomination Award. He has authorized and co-authorized on more than 80 papers in top-tier academic journals and conferences. He serves as an area chair for the conference of VISAPP. He has been serving as a reviewer for numerous academic journals and conferences such as TPAMI, IJCV, TIP, TMM, TCYB, CVPR, ICCV, ECCV and NeurIPS.
\end{IEEEbiography}

\begin{IEEEbiography}[{\includegraphics[width=1in,height=1.25in,clip,keepaspectratio]{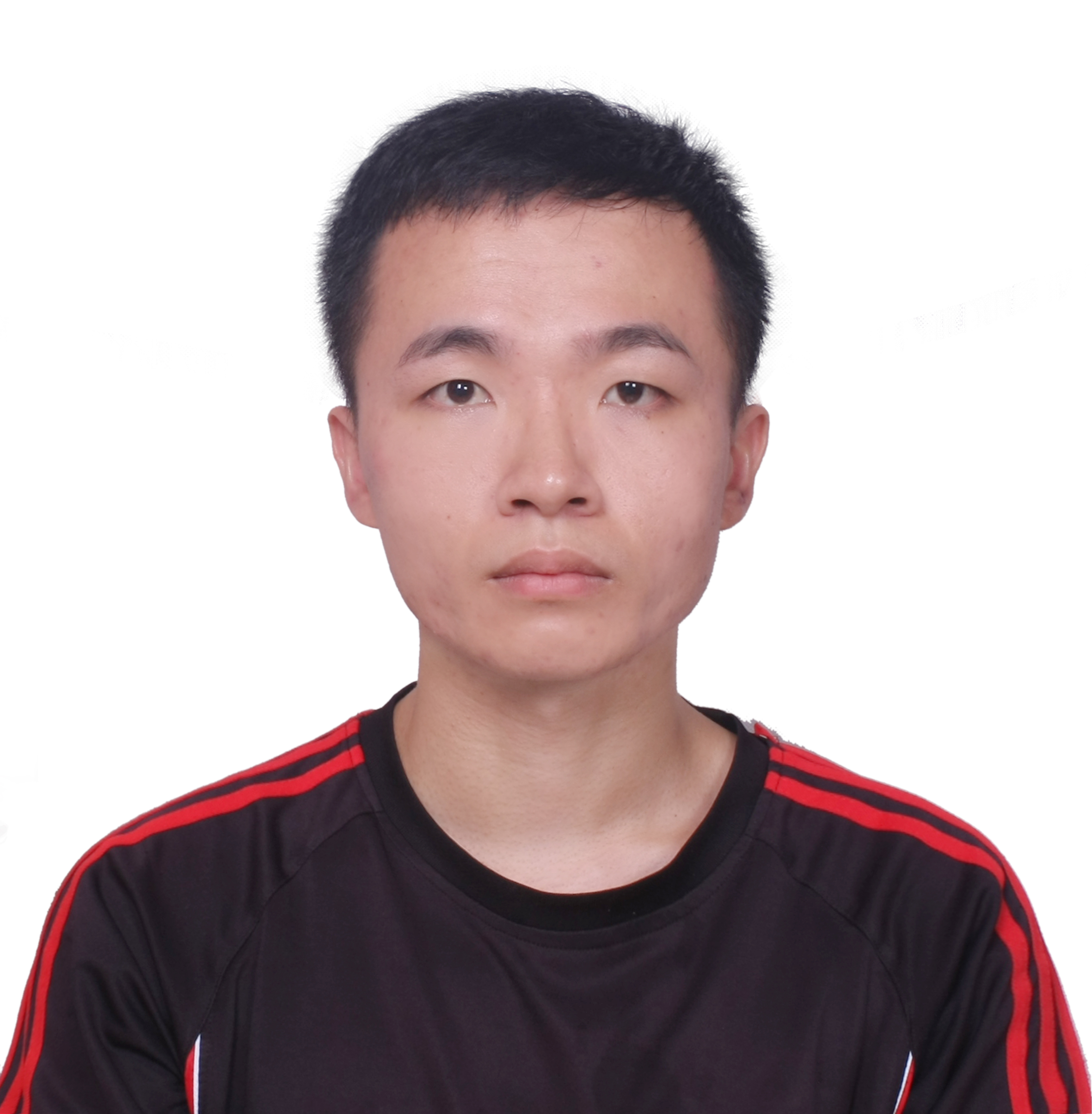}}]{Ziyi Wu}
received the B.E. degree from the School of Computer Science and Engineering, Sun Yat-sen University, Guangzhou, China, in 2020, where he is currently pursuing the Master's degree in computer science. His current research interests include salient object detection and self-supervised learning.
\end{IEEEbiography}

\begin{IEEEbiography}[{\includegraphics[width=1in,height=1.25in,clip,keepaspectratio]{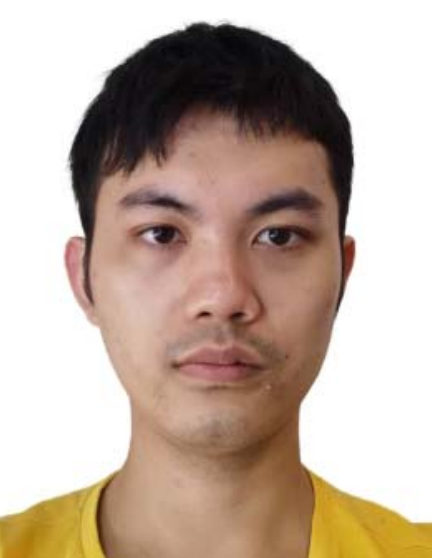}}]{Junfan Lin}
received the B.S. degree in software engineering from Sun Yat-sen University, Guangzhou, China, in 2017, where he is currently pursuing the Ph.D. degree in computer science and technology, advised by Prof. Liang Lin. His research interests include reinforcement learning and natural language processing.
\end{IEEEbiography}

\begin{IEEEbiography}[{\includegraphics[width=1in,height=1.25in,clip,keepaspectratio]{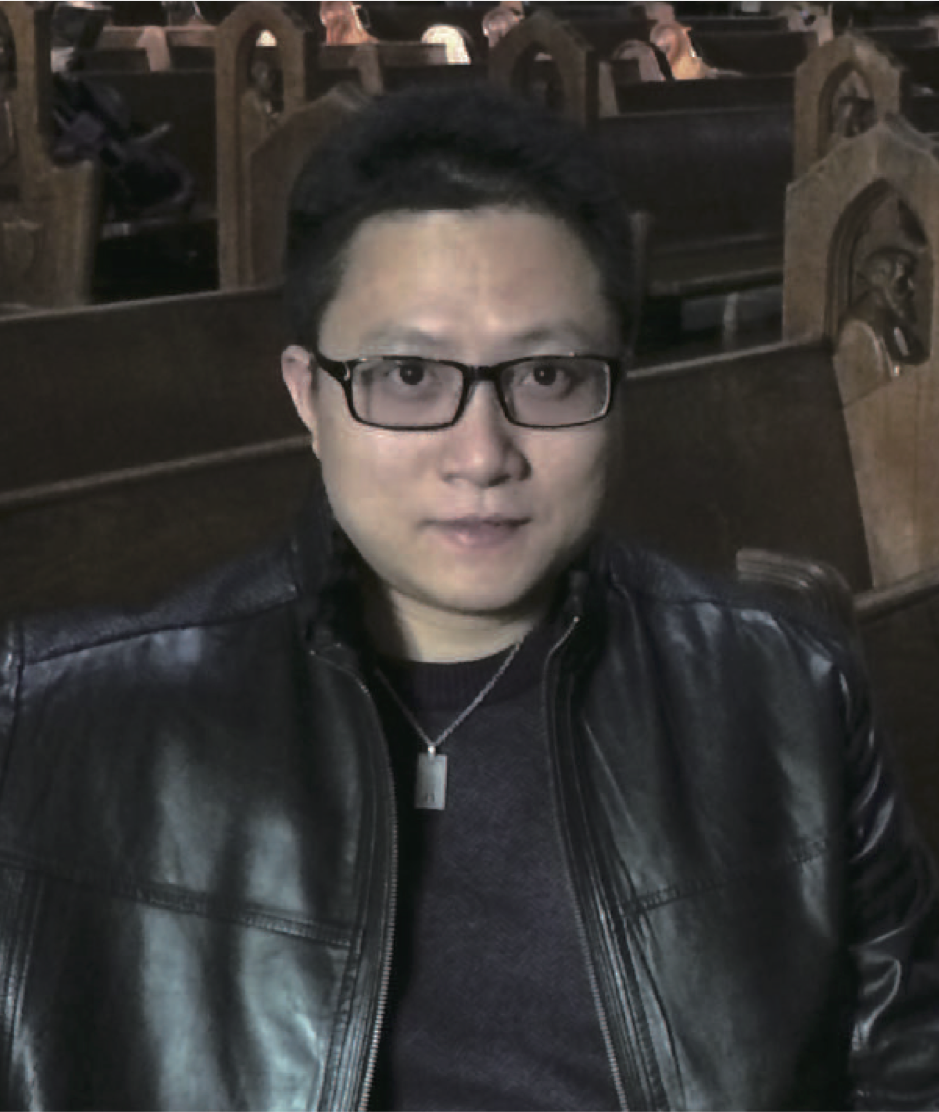}}]{Liang Lin}
is a full Professor of Sun Yat-sen University. He is the Excellent Young Scientist of the National Natural Science Foundation of China. From 2008 to 2010, he was a Post-Doctoral Fellow at University of California, Los Angeles. From 2014 to 2015, as a senior visiting scholar, he was with The Hong Kong Polytechnic University and The Chinese University of Hong Kong. From 2017 to 2018, he leaded the SenseTime R\&D teams to develop cutting-edges and deliverable solutions on computer vision, data analysis and mining, and intelligent robotic systems. He has authorized and co-authorized on more than 100 papers in top-tier academic journals and conferences. He has been serving as an associate editor of IEEE Trans. on Neural Networks and Learning Systems, IEEE Trans. Human-Machine Systems, The Visual Computer and Neurocomputing. He served as Area/Session Chairs for numerous conferences such as ICME, ACCV, ICMR. He was the recipient of Best Paper Nomination Award in ICCV 2019, Best Paper Runners-Up Award in ACM NPAR 2010, Google Faculty Award in 2012, Best Paper Diamond Award in IEEE ICME 2017, and Hong Kong Scholars Award in 2014. He is a Fellow of IET.
\end{IEEEbiography}

\end{document}